\documentclass{article}

\usepackage{microtype}
\usepackage{multicol}
\usepackage{enumitem} 

\usepackage{graphicx}
\usepackage{subcaption}
\usepackage{multirow}
\usepackage{booktabs} 

\usepackage{mdframed}  
\usepackage{caption}   

\newcounter{examplecounter}
\renewcommand{\theexamplecounter}{\arabic{examplecounter}}  

\newenvironment{example}[2][] 
    {\refstepcounter{examplecounter}%
    \begin{mdframed}[backgroundcolor=gray!10, linewidth=1pt, skipabove=10pt, innerleftmargin=10pt, innerrightmargin=10pt, innertopmargin=10pt, innerbottommargin=10pt]%
    \noindent\textbf{\small Example~\theexamplecounter: \ #1}\label{#2}\par\smallskip}%
    {\end{mdframed}}

\usepackage[accepted]{icml2025}

\usepackage{amsmath}
\usepackage{amssymb}
\usepackage{mathtools}
\usepackage{amsthm}
\usepackage{xcolor}         
\usepackage{placeins}

\definecolor{royalazure}{rgb}{0.0, 0.22, 0.86}
\definecolor{royalblue}{rgb}{0.04,0.33,0.64 }
\usepackage[
    colorlinks,
    citecolor=royalazure,
    filecolor=royalazure,
    linkcolor=royalazure,
    urlcolor=blue
]{hyperref}
\usepackage[normalem]{ulem}
\usepackage{amsmath}
\usepackage{algpseudocode}
\usepackage{amsfonts}
\usepackage{array}

\usepackage{enumitem}
\usepackage{tabularx} 
\usepackage{booktabs}
\usepackage{colortbl} 
\usepackage{pifont} 
\usepackage[separate-uncertainty = true,retain-zero-uncertainty=true]{siunitx} 

\newcommand{\cmark}{\ding{51}} 
\newcommand{\xmark}{\ding{55}} 

\usepackage[capitalize,noabbrev]{cleveref}

\theoremstyle{plain}

\theoremstyle{definition}

\theoremstyle{remark}

\usepackage[acronym]{glossaries}
\glsdisablehyper

\newacronym{ce}{CE}{cross-entropy}
\newacronym{ntl}{NTL}{Number Token Loss}

\newcommand{\acs}{\acrshort}
\newcommand{\acf}{\acrfull}

\usepackage[textsize=tiny]{todonotes}
\usepackage{caption}
\usepackage[hang,flushmargin]{footmisc}
\usepackage[font=small]{caption}

\icmltitlerunning{NTL -- The Number Token Loss}

\begin{document}

\twocolumn[
\icmltitle{
Regress, Don't Guess --\\ A Regression-like Loss on Number Tokens for Language Models
}



\icmlsetsymbol{equal}{*}

\begin{icmlauthorlist}
\icmlauthor{Jonas Zausinger}{equal,TUM_ai,TUM}
\icmlauthor{Lars Pennig}{equal,TUM_ai,TUM,Helmholtz,MCML}
\icmlauthor{Anamarija Kozina}{TUM_ai,TUM}
\icmlauthor{Sean Sdahl}{TUM_ai,TUM}
\icmlauthor{Julian Sikora}{TUM_ai,TUM}
\icmlauthor{Adrian Dendorfer}{TUM_ai,TUM}
\icmlauthor{Timofey Kuznetsov}{TUM_ai,TUM}
\icmlauthor{Mohamad Hagog}{TUM_ai,LMU}
\icmlauthor{Nina Wiedemann}{ETH}
\icmlauthor{Kacper Chlodny}{TUM_ai,TUM}
\icmlauthor{Vincent Limbach}{TUM_ai,TUM}
\icmlauthor{Anna Ketteler}{TUM_ai,TUM}
\icmlauthor{Thorben Prein}{TUM_ai,TUM}
\icmlauthor{Vishwa Mohan Singh}{TUM_ai,LMU}
\icmlauthor{Michael Danziger}{IBM}
\icmlauthor{Jannis Born}{IBM}

\end{icmlauthorlist}

\icmlaffiliation{TUM}{Technical University of Munich}
\icmlaffiliation{Helmholtz}{Helmholtz Munich}
\icmlaffiliation{MCML}{Munich Center for Machine Learning (MCML)}
\icmlaffiliation{LMU}{LMU Munich}
\icmlaffiliation{TUM_ai}{TUM.ai}
\icmlaffiliation{IBM}{IBM Research Europe}
\icmlaffiliation{ETH}{ETH Zurich}

\icmlcorrespondingauthor{Jannis Born}{jab@zurich.ibm.com}

\icmlkeywords{Machine Learning, ICML, Language Models, Mathematical Reasoning, LLMs, Loss Functions}

\vskip 0.3in
]



\printAffiliationsAndNotice{\icmlEqualContribution} 

\begin{abstract}
  While language models have exceptional capabilities at text generation, they lack a natural inductive bias for emitting numbers and thus struggle in tasks involving quantitative reasoning, especially arithmetic.
  One fundamental limitation is the nature of the \acf{ce} loss, which assumes a nominal scale and thus cannot convey proximity between generated number tokens.
  In response, we here present a regression-like loss that operates purely on token level. 
  Our proposed \acf{ntl} comes in two flavors and minimizes either the $L_p$ norm or the Wasserstein distance between the \textit{numerical values} of the real and predicted number tokens.
  NTL can easily be added to any language model and extend the \acs{ce} objective during training without runtime overhead.
  We evaluate the proposed scheme on various mathematical datasets and find that it consistently improves performance in math-related tasks.
  In a direct comparison on a regression task, we find that NTL can match the performance of a regression head, despite operating on token level.
  Finally, we scale NTL up to $3$B parameter models and observe improved performance, demonstrating its potential for seamless integration into LLMs.
  We hope to inspire LLM developers to improve their pretraining objectives and distribute NTL as a minimalistic and lightweight PyPI package \texttt{ntloss}: 
    {\urlstyle{same}  \href{https://ibm.biz/ntl-pypi-repo}{https://github.com/ai4sd/number-token-loss}}.
  Development code for full paper reproduction is  available separately.

\end{abstract}

\begin{figure}[htb!]

    \begin{subfigure}[b]{\linewidth}
        \centering
        \includegraphics[width=0.95\linewidth]{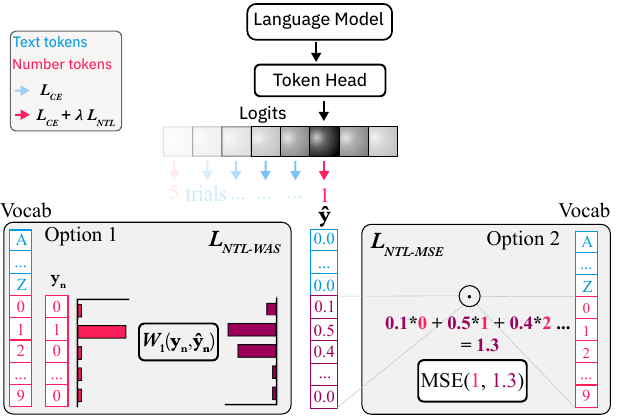}
        \caption{The two flavors of the proposed Number Token Loss (NTL).}
        \label{fig:xval-ntl}
    \end{subfigure}

        \vspace{1em} 

    \centering
    \begin{subfigure}[b]{\linewidth}
        \centering
        \includegraphics[width=0.85\linewidth]{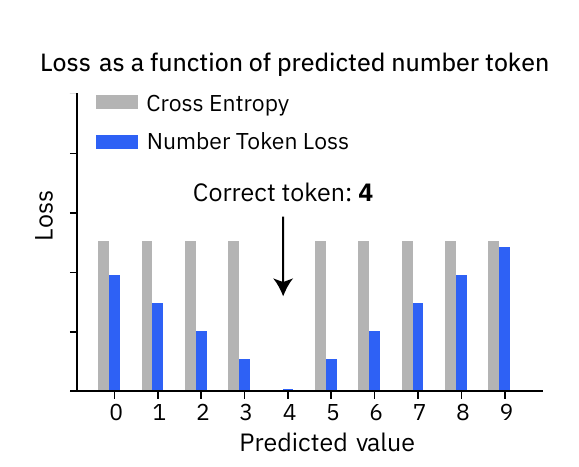}
        \caption{Comparison of NTL to CE.}
        \label{fig:plot1}
    \end{subfigure}

    \caption{\textbf{(a) \acf{ntl}.} The NTL allows computing a regression-like loss directly on a language modeling head.
        We propose two schemes: $\mathcal{L}$\textsubscript{NTL-MSE} uses a dot product of number token values and their probabilities;
        $\mathcal{L}$\textsubscript{NTL-WAS} uses the Wasserstein-1 distance between label and prediction distributions. 
        \textbf{(b)}
        \acs{ce} is nominal-scale and thus assigns equal loss to all incorrect predictions whereas NTL increases with distance from ground truth like a regression loss.
        }
    \label{fig:combined}
\end{figure}


\section{Introduction}

As coined by~\citet{thawani2021representing}, numbers in natural texts are \textit{ubiquitous} and \textit{important}, yet systematically \textit{neglected} by language models (LMs).
Even worse, while Transformers~\cite{vaswani2017attention} were invented for NLP, they have permeated various scientific domains~\cite{jumper2021highly,boiko2023autonomous}, where tabular/numerical data is more prevalent than in NLP and often governs task definitions: Molecules are labeled with drug efficacy, chemical reactions with yield, and synthesis procedures are natural text interspersed with quantities and times.
%
Prior art for joint language-number modeling suggested the use of verifiers~\cite{cobbe2021training,li2023making}, 
chain-of-thought (CoT) reasoning~\cite{zhong2024achieving, wei2022chain, lee2023recursion}, calculators or code interpreters~\cite{qu2025tool, gao2023pal} to yield improved performance in LMs. 
On a model-level, customized ideas such as reverting numbers~\cite{zhang2024reverse}, right-to-left tokenization~\cite{singh2024tokenizationcountsimpacttokenization, lee2023teachingarithmeticsmalltransformers}, or joint prediction of multiple number tokens have been proposed~\cite{bachmann2024pitfallsnexttokenprediction}. 
Still, LMs notoriously struggle even with simple arithmetic tasks like three-digit multiplication~\cite{dziri2024faith}.
We argue that all these strategies  
-- reformulating the task, trying to correct answers \textit{a posteriori} with calculators, or using significantly more compute (CoR) -- avoid the fundamental, underlying problem: 
number representation in LMs is poor, for multiple reasons:
\begin{enumerate}[leftmargin=*, topsep=0pt, partopsep=0pt, itemsep=0pt, parsep=0pt]
    \item \textbf{Tokenization}: Standard subword tokenization splits numbers into arbitrary tokens, disrupting their structure and losing a significant amount of numerical information \cite{wallace2024DoNLPKnowNumbers}. Mitigation strategies include scientific notation~\cite{zhang-etal-2020-language-embeddings} or digit-level tokenization~\cite{geva2020injecting}, which may also preserve the decimal order of each digit~\cite{born2023regression}.
    \item \textbf{Embedding}: Canonically, LMs have to recover the structure of numbers from data because number token embeddings are learned like any other.
    To mitigate that, many flavors of numeracy-preserving word embeddings exist~\cite{sundararaman2020methods,born2023regression,golkar2023xval}, often akin to positional encodings and sometimes adapted for special cases like angular embeddings for modular arithmetics~\cite{stevens2024salsafrescaangularembeddings,saxena2024teachingtransformersmodulararithmetic}.
    \item \textbf{Sequential prediction:} Token-wise decoding fails to account for the greater significance of higher-order digits in a number. \citet{zhang2024reverse} find that reversing the digit order and human-inspired, step-by-step calculations for addition and multiplication improve accuracy. 
    \item \textbf{Training objective}: Cross-entropy (CE) loss assumes a nominal scale, thus it fails to convey the proximity between numbers, effectively inducing a semi-supervised setting. For example, predicting a \texttt{3} instead of a \texttt{2} will not generally induce lower loss than a \texttt{9}.
\end{enumerate}
%
In this paper, we aim to equip LMs with better inductive biases to handle combinations of textual and numerical data, such as math word problems or scientific datasets.
We propose two versions of a regression loss on number tokens that respect numerical proximity and can be effectively combined with standard \acs{ce} (\autoref{fig:xval-ntl}).
The first version of the number token loss (NTL) computes the Mean Squared Error (MSE) between the numerical value of the label and the predicted class probabilities weighted by their numerical token value.
The second version computes the Wasserstein distance between the distribution of the predicted number probabilities and the ground truth distribution, which is the one-hot encoding of the label.


In both cases, the NTL has the following key characteristics. 
\textbf{(1) NTL is model agnostic}. 
It can be applied to any language model (Transformer, Mamba, etc.) in any style (encoder-decoder, decoder-only etc). 
\textbf{(2) NTL is plug-and-play for any LM} because it makes minimal assumptions about the vocabulary.
It only requires a map between tokens (strings) and their numerical value (float) and is thus compatible with digit-level as well as multi-digit tokenization\footnote{digit-level, e.g.: LLaMA2~\cite{touvron2023llama2openfoundation}, DeepSeek-V2~\cite{deepseekai2024deepseekv2strongeconomicalefficient} and PaLM~\cite{chowdhery2022palmscalinglanguagemodeling}; multi-digit, e.g. LLaMA3~\cite{grattafiori2024llama3herdmodels} or T5~\cite{raffel2020exploring}.  
}.
\textbf{(3) NTL does not add computational overhead}. 
The most capable NTL version (NTL-WAS) slows down the loss calculation only by 1\% compared to cross-entropy.
This difference becomes negligible over the whole training step (see later in~\autoref{fig:Link result}).

Our empirical analysis further shows that:
\textbf{(4) NTL consistently improves performance in mathematical tasks} across architectures, compared to cross-entropy alone.
\textbf{(5)} On a real regression task, \textbf{a LM with NTL performs on-par with a regression head} and improves by 10\% over a LM with standard \acs{ce}.
\textbf{(6) NTL does not hamper performance on text-only tasks} and (7) \textbf{NTL scales well to billion-size models}.
Therefore nothing speaks against using it during LLM pretraining.

\section{Number Token Loss (NTL)}

Inevitably, some tokens of any LM correspond to digits or numbers, not to text.
Our basic idea is to combine the logits of these \textit{number tokens} with their numerical values to compute a loss.
Such a loss can effectively augment cross-entropy by taking numerical proximity of number tokens into account.
Note that outside NLP, some systematic studies of "regression by classification" exist~\cite{shah2023learning,stewart2023regression}, yet a principled approach that facilitates seamless integration into LLMs lacks entirely.

\begin{figure}[htb!]
    \hfill
        \centering
        \includegraphics[width=\columnwidth]{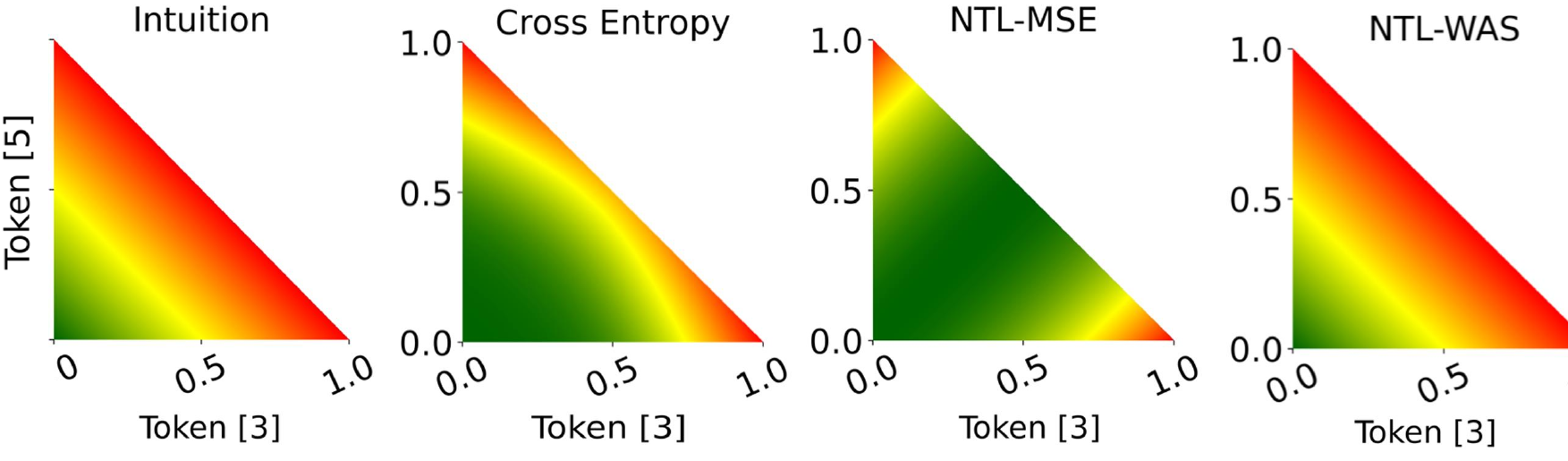}
        \vspace{-8mm}
        \caption{
        The heatmap plot shows the respective loss for a given combination of the class probabilities for tokens $3$ and $5$, where the ground truth is token $4$. The behavior of the NTL-WAS is closest to the intuitive desired behavior of the loss function, while the NTL-MSE does not have a unique minimum. 
        }
        \label{fig:plot2}
\end{figure}

\paragraph{NTL with $\mathcal{L}_p$ norm (e.g., NTL-MSE)}
This loss maps the predicted token probabilities to a real-valued output by calculating the dot product of the probabilities and their corresponding numerical token value (cf.~\autoref{fig:xval-ntl}). This transformation enables the application of standard regression losses, such as the MSE, to compare the predicted output against the numerical value of the ground truth token.
Assume a model $f(\cdot)$, input tokens $\mathbf{x}_{\leq i}$ ($i \leq N$) and a vocabulary $V$.
Let $\hat{\mathbf{y}}_i:=f(\mathbf{x}_{\leq i})$ be the predicted probability distribution, $\mathbf{y}_i$ the one-hot label with $y_i$ as the \textit{numerical} value of the ground truth token and $\mathcal{V}: V \rightarrow \mathbb{R}$ a map to convert tokens (strings) to their numerical values (floats), with the index range $s...t$ representing the number tokens. 
Then we compute NTL-MSE:
\begin{equation}
    \mathcal{L}_{\text{NTL-MSE}} = \frac{1}{N}\sum_{i}^N (y_i - \mathbf{\hat{y}}_{i}^{s:t} \circ \mathcal{V}^{s:t} )^2 
    \label{eq:NTL-MSE}
\end{equation}
Notably, instead of a nominal-scale loss like \acs{ce}, the NTL-MSE effectively conveys proximity between numbers. 
For example, if the label is \texttt{4} and the LM predicts \texttt{5} instead of \texttt{9}, the loss will be lower, matching our intuitive expectation.
This is in contrast to the \acs{ce} which gives constant loss regardless of the proximity of the number due to its nominal nature (cf.~\autoref{fig:plot1}).
By changing the $p$-order in NTL-MSE, different $L_p$-norm losses can be obtained (e.g., NTL-MAE). Huber loss is also compatible.
While NTL-$L_p$ describes a family of losses, we note that NTL-MSE has been proposed already in concurrent work by~\citet{lukasik2025better} where it is called RAFT and has shown improved performance of LLMs on pure regression tasks with up to five number digits.
%
%
However, NTL-$L_p$ has a non-unique minimum and can thus return spuriously low loss for incorrect predictions.
Consider e.g., a sample with label \texttt{4} where $50\%$ of the mass is on \texttt{0} and $50\%$ on token \texttt{8}, then NTL will be zero (see \autoref{fig:plot2}).
While such cases are rare due to the softmax emphasizing logit differences, combining NTL with \acs{ce} loss helps correct spurious cases.
But to address the non-unique minima more systematically we propose a second, refined version.

\FloatBarrier
\paragraph{NTL with Wasserstein-1 Distance (NTL-WAS)}
To measure the similarity of a predicted probability distribution of number tokens and a ground truth distribution, we take inspiration from Optimal Transport (OT) and leverage the discrete Wasserstein-1 distance, generally defined as:
\begin{equation}
 W_c(P, Q) = \min_{\gamma\in \Gamma(P, Q)} \sum_i \sum_j \gamma_{ij} c(u_i - v_j)
 \label{general_wasserstein}
\end{equation}
where $u_i$ and $v_j$ are points, $P$ and $Q$ their associated probabilities respectively, and $c$ a function specifying the transport costs. $\gamma$ is a coupling between $P$ and $Q$, where $\gamma_{ij}$ specifies the mass to be transported from $u_i$ to $v_j$. 
We apply OT to measure the difference between the distributions $\mathbf{\hat{y}_i}$ and $\mathbf{y_i}$ by defining the cost function $c$ as the Euclidean distance between the numerical token values:
\begin{equation}
W_{\text{LM}}(P, Q) = \min_{\gamma\in \Gamma(\mathbf{y}_i^{s:t}, \mathbf{\hat{y}}_i^{s:t} )} \sum_{j=s}^{t} \sum_{k=s}^{t} \gamma_{jk} \Vert \mathcal{V}^j - \mathcal{V}^k \Vert
\label{eq:value_diff_wasserstein}
\end{equation}

\paragraph{Note on Cost Function:}
~\autoref{eq:value_diff_wasserstein} is very flexible -- by changing the cost function $c$, NTL can be applied even when the number tokens lie in arbitrary, non-Euclidean spaces. 
The cost function does not need to adhere to the definition of a distance metric and can instead be specified as a pairwise matrix between number tokens. 
NTL could thus find utility also in exotic cases such as modular arithmetics~\cite{charton23learning,kera2024learning,gromov2023grokking}. 
In this paper, we mainly assume standard Euclidean distance due to its general applicability.
A practical example where a non-euclidean cost is useful are multi-digit tokenizers which often contain individual tokens for very large number (Section~\ref{SEC:Tokenizers} contains explicit results on non-euclidean cost).

Computing the quantity in~\autoref{eq:value_diff_wasserstein} requires solving a linear program and is thus not generally differentiable. 
Approximations could be made for the general case via entropic regularization~\cite{cuturi2013sinkhorn}, but our application of $W_c$ entails two special cases that enable its efficient and differentiable computation: First, if $\mathbf{y}_i$ is one-hot, $\mathcal{L}_{\text{NTL-WAS}}$ coincides with the weighted sum of the absolute differences from the label and the numerical token values:
\begin{equation} 
    \mathcal{L}_{\text{NTL-WAS}} = \frac{1}{N} \sum_{i=1}^{N} \sum_{j=s}^t \hat{\mathbf{y}}_i^j |y_i - \mathcal{V}^j|
    \label{eq:ntl-was}
\end{equation}

Second, if the number token indices $s...t$ are sorted by numerical value $\mathcal{V}^{s:t}$ and the values are equidistantly spaced, the Wasserstein-1-distance can be computed using the cumulative distribution function $\mathrm{CDF}(\cdot)$: 
\begin{equation} 
    \mathcal{L}_{\text{NTL-WAS-CDF}} = \frac{1}{N} \sum_{i=1}^{N} \left| \mathrm{CDF}\left( {\mathbf{y}}_i^{s:t} \right) - \mathrm{CDF}\left( \mathbf{\hat{y}}_i^{s:t} \right) \right|
    \label{eq:ntl-was-ot}
\end{equation}
The advantage of $\mathcal{L}_{\text{NTL-WAS-CDF}}$  is that it does not require the label $\mathbf{y}_i$ to be one-hot -- instead it supports arbitrary target distributions obtained e.g., via label smoothing on number tokens~\cite{wang2025tokenmol}.
In practice, unless mentioned otherwise, we compute $\mathcal{L}_{\text{NTL-WAS}}$ via~\autoref{eq:ntl-was} because it is $230$x faster than~\autoref{eq:ntl-was-ot} and does not require sorting or equidistant spacing and is thus applicable beyond digit-level tokenization.
As one can see in \autoref{fig:plot1}, $\mathcal{L}_{\text{NTL-WAS}}$ not only conveys proximity between numbers correctly but also eliminates the problem of non-unique minima in $\mathcal{L}_{\text{NTL-MSE}}$~(\autoref{fig:plot2}).
Both versions of the NTL are scaled with $\lambda$ (default $0.3$) and added to the standard \acs{ce} loss:
\begin{equation}
    \mathcal{L} = \mathcal{L}_{CE} + \lambda \mathcal{L}_{NTL}
    \label{eq:loss}
\end{equation}
Note that both versions of the NTL yield zero loss for all non-numerical tokens (see Appendix~\ref{algo:ntl} for pseudo-code).

\FloatBarrier




\section{Experiment Setup}
\subsection{Backbone T5 and Model Variants}
We use a T5 backbone~\cite{raffel2020exploring} (Appendix~\ref{Appendix:T5}) for most of our experiments, unless stated otherwise, due to its flexible encoder-decoder architecture and its success in various natural language processing tasks. We built upon the T5 implementation and language modeling trainer from \texttt{transformers}~\cite{wolf2020transformers}.
For the multitask mathematics experiments in Section \ref{results-mathematic-dataset} we use T5-Base with $220$M parameters.
For the scaling experiment in Section \ref{results-gsm8k-dataset} we use T5-3B. 
For ablation studies and other experiments, we use T5-Small with $60$M parameters. 
To illustrate the model-agnostic nature of NTL, we evaluate it on both GPT-2 and IBM Granite model variants.
We always initialize the models with their pre-trained weights.
Building up on prior work~\cite{geva2020injecting}, we adapt the tokenization scheme to tokenize all numbers on the digit level, while leaving all other tokens unchanged. Although both NTL versions are compatible with any tokenizer, single-digit tokenization improves performance for both \acs{ce} and NTL on mathematical tasks in our experiments (see~\autoref{SEC:Tokenizers}).



\subsection{Baselines}
We compare to three related methods.

First, the \textbf{Regression Transformer}~\cite{born2023regression} which tokenizes numbers on digit level (considering both the position and value of each digit) 
and leverages a fixed number embedding that preserves relative proximity of the numbers (details in Appendix~\ref{Appendix:RT}).

Second, \textbf{xVal}~\cite{golkar2023xval} which encodes real numbers using a single \texttt{[NUM]} token multiplied by their numerical value. 
For decoding, a number/regression head predicts the value while the token head outputs the sequence, replacing \texttt{[NUM]} during inference.
However, this scheme is incompatible with T5  (see Appendix~\ref{Appendix:xVal}) and problematic for beam search decoding. 
We thus use the xVal encoder and masked language modeling in our experiments. 
Third, \textbf{Gaussian Cross Entropy} (or Gaussian label smoothing;~\citet{wang2025tokenmol}) which
addresses the limitations of standard \acs{ce} for continuous-valued tokens.
Instead of one-hot encoded labels, a Gaussian distribution is centered on the true numeric label. In this scheme, nearby numeric tokens receive partial probability mass, reflecting the continuous nature of numerical values. 
Given a numeric label $y_{i}$,  
its one-hot vector $\mathbf{y_i}$ is replaced with a Gaussian distribution:
    \begin{equation}
        \mathbf{\tilde{y}}^j_i \;=\; \frac{1}{\sqrt{2\pi\,\sigma^2}}
        \exp\!\Bigl(-\tfrac{(\mathcal{V}^j-y_{i})^2}{2\sigma^2}\Bigr),\ \ \ \forall j\in [s..t]
    \end{equation}
where $\sigma$ governs the smoothing width. 
The objective remains the cross-entropy, but now between the modified labels $\mathbf{\tilde{y}}_i$ and the predicted distribution $\hat{\mathbf{y}}_i$:
    \begin{equation}
        \mathcal{L}_{\text{GCE}} \;=\; -\frac{1}{N}\sum_{i=1}^N 
        \sum_{j=s}^t \mathbf{\tilde{y}}^j_i \,\log \mathbf{\hat{y}}^j_i ,
    \end{equation}
The obvious disadvantage is that the smoothing is entirely fabricated: there is no real uncertainty in the label.
While this technically provides incorrect labels, it ensures that the inherent similarity among nearby numeric tokens is captured and continuity in numeric space is preserved.
\citet{wang2025tokenmol} finds that this does not alter classification performance on strictly categorical tokens.


\section{Experimental Results}
Across all experiments, we always optimize NTL jointly with CE (cf.~\autoref{eq:loss}; default $\lambda=0.3$).
For brevity we refer to this as just NTL-MSE or NTL-WAS.
\subsection{NTL Improves Performance in Arithmetics}\label{results-mathematic-dataset}

To test the mathematical capabilities of our models, we use more than 25 million samples from the mathematical Q\&A dataset from DeepMind~\cite{saxton2019analysingmathematicalreasoningabilities}.
The dataset comes with two sets of tests: interpolation tests, one for each type of question occurring in the training set, and extrapolation tests, which measure generalization along various axes of difficulty beyond that seen during training.
We provide more information about the dataset in Appendix~\ref{dataset-details}.

\subsubsection{Multitask mathematics dataset}

We evaluated all models on the two test sets of this dataset
and report the accuracy (the proportion of exactly predicted numbers), as well as the Mean Absolute Error (MAE) and the R\textsuperscript{2}-score.
Since the dataset is skewed with some very high values, we perform a $log_{10}$ transformation on the predicted and ground truth numbers before calculating MAE and R\textsuperscript{2}-score. 
All models were trained for $\sim 1$M steps with a batch size of 32.
More details on the models’ training hyperparameters can be found in Appendix~\ref{app:hyperparameters}.

\begin{table}[htb!]
\centering
\caption{\textbf{Evaluation on Mathematics Dataset}}
\label{tab:table_result}
    \centering
    \caption*{(a) Interpolated Test Data}
    \vspace{1em}
    \resizebox{0.83\linewidth}{!}{
    \begin{tabular}{@{}llccc@{}}
        \toprule
        \textbf{Model}            &
        \textbf{Loss}            &\textbf{Accuracy}   & \textbf{MAE}  & \textbf{R\textsuperscript{2}} \\
        \midrule
        T5   & CE            & 0.64          & 0.13         & 0.97                        \\
        T5 & \textbf{NTL-MSE}   & 0.72 & 0.11 & 0.97 \\
        T5 & \textbf{NTL-WAS}   & \textbf{0.75} & \textbf{0.10}& \textbf{0.98}               \\
        T5 & \textbf{NTL-MAE}   & 0.74 & \textbf{0.10} & 0.97 \\
        
        RT  & CE                      & 0.71          & 0.11         & 0.97                        \\
        xVal   & MSE                   & 0.10         & 0.26         & 0.97                        \\
        \bottomrule
    \end{tabular}
    }
    \label{tab:table_interpolation}
\hfill 
    \centering
    \caption*{(b) Extrapolated Test Data}
    \vspace{1em}
    \resizebox{0.83\linewidth}{!}{
    \begin{tabular}{@{}llccc@{}}
        \toprule
        \textbf{Model}            & 
        \textbf{Loss}      &
        \textbf{Accuracy}   & \textbf{MAE}  & \textbf{R\textsuperscript{2}} \\
        \midrule
        T5 & CE               & 0.367          & 0.785        & \textbf{0.913}            \\
        T5 & \textbf{NTL-MSE}   & 0.428          & 0.779 & 0.909                       \\
         T5 &\textbf{NTL-WAS}   & \textbf{0.432}          & \textbf{0.744}& \textbf{0.913}                  \\
         T5 & \textbf{NTL-MAE}   & 0.427 & 0.792 & 0.906 \\
        RT     & CE                   & 0.404          & 0.987        & 0.738                        \\
        xVal    & MSE                  & 0.058          & 0.826        & 0.819                        \\
        \bottomrule
    \end{tabular}
    }
    \label{tab:table_extrapolation}
\normalsize 
\end{table}

The results in~\autoref{tab:table_result} and~\autoref{fig:res-bar-comparison} show that vanilla T5 clearly benefits from both NTL variants.
In fact, the accuracy increases by more than $10\%$ for NTL-WAS compared to vanilla T5 in the interpolation tasks.
The NTL-WAS was found to have the best performance across all three metrics and both interpolation and extrapolation tasks.
Moreover, NTL consistently outperforms competing methods such as the RT and xVal.
This confirms our hypothesis that number representation in LMs can be effectively improved through a minor, architecture-agnostic loss modification.

Since xVal is equipped with a regression head, numbers are never predicted exactly, so the accuracy value for xVal refers to the predictions rounded to two decimal places.
The poor performance of xVal can likely be explained by the extensive range of numbers in the used dataset.
The effective number range of xVal is limited to $[-5, 5]$ due to the combination of its scaling of the number token embeddings and the pre-layer-norm in the backbone~\cite{xiong2020layer,golkar2023xval}.
To take this into account, we scale our dataset for xVal with $log(1+x)$. However, this means that large numbers can no longer be adequately distinguished by the model, as their embeddings become very similar.
For a direct comparison without any modifications to the xVal processing, we repeated the 3-digit multiplication experiment from \citet{golkar2023xval}. Again, our method beats xVal. Details can be found in Appendix \ref{Appendix:xVal}.


\subsubsection{Ablation studies}

In order to investigate the impact of variations in loss functions, we conducted extensive experiments testing nine different configurations: standard cross entropy loss, NTL-MSE and NTL-WAS (both with $\lambda \in \{0.3, 0.8, 2\}$), NTL-MAE and NTL-Huber. 
Training of these models was done on a subtask of the original dataset (100k samples), 
consisting of basic integer arithmetic Q\&A pairs, where the input is a short natural language question containing an arithmetic expression, and the output is a single integer. 
A comparison of their performances on the interpolation and extrapolation test sets is shown in~\autoref{tab:ablstudies_both}. 

\begin{table}[htb!]
\centering
\caption{
\textbf{Evaluation of NTL Loss Variants
}}
    \centering
    \caption*{(a) Interpolated Test Data}
    \vspace{1em}
    \begin{minipage}{0.48\textwidth}
    \resizebox{\linewidth}{!}{
    \begin{tabular}{lcccc}
    \toprule
     \textbf{Loss} & $\mathbf{\lambda}$ &          \textbf{Accuracy} &               \textbf{MAE} &                \textbf{R\textsuperscript{2}} \\
\midrule
         CE &        & $0.34_{\pm 0.01}$ & $2.15_{\pm 0.08}$ & $0.95_{\pm 0.01}$ \\
         \midrule
    \multirow{4}*{NTL-MSE} 
     &    0.15 &$0.44 _{\pm 0.02}$ & $0.92 _{\pm 0.05}$ & $0.99 _{\pm 0.00}$ \\
     &    0.3 & $0.41_{\pm 0.01}$ & $1.02_{\pm 0.06}$ & $0.99_{\pm 0.00}$ \\
     &    0.8 & $0.37_{\pm 0.02}$ & $1.29_{\pm 0.11}$ & $0.99_{\pm 0.00}$ \\
     &    2.0 & $0.33_{\pm 0.01}$ & $1.67_{\pm 0.20}$ & $0.97_{\pm 0.01}$ \\
    \midrule 
\multirow{4}*{NTL-WAS} 
& 0.15 & $0.44 _{\pm 0.02}$ & $0.93 _{\pm 0.01}$  & $0.99 _{\pm 0.00}$ \\
&    0.3 & $0.43_{\pm 0.05}$ & $0.91_{\pm 0.06}$ & $0.99_{\pm 0.00}$ \\
&    0.8 & $0.43_{\pm 0.04}$ & $0.94_{\pm 0.07}$ & $0.99_{\pm 0.00}$ \\
 &    2.0 & $0.41_{\pm 0.06}$ & $1.01_{\pm 0.08}$ & $0.99_{\pm 0.00}$ \\
 \midrule
NTL-Huber &    0.3 & $0.44_{\pm 0.02}$ & $\textbf{0.89}_{\pm 0.03}$ & $\textbf{1.00}_{\pm 0.00}$ \\
\midrule
NTL-MAE &    0.3 & $\textbf{0.45}_{\pm 0.02}$ & $\textbf{0.89}_{\pm 0.07}$ & $0.99_{\pm 0.00}$ \\
\bottomrule
\end{tabular}
    }
    \label{tab:ablstudies_interpolate}
\end{minipage}
\vspace{1em}
\begin{minipage}{0.48\textwidth}

\centering
    \centering
    \caption*{(b) Extrapolated Test Data}
    \vspace{1em}
    \resizebox{\linewidth}{!}{

\begin{tabular}{lcccc}
\toprule
     \textbf{Loss} & $\mathbf{\lambda}$ & \textbf{Accuracy} &                \textbf{MAE} &                \textbf{R\textsuperscript{2}} \\
\midrule
CE &        & $0.05_{\pm 0.00}$ & $61.91_{\pm 1.31}$ & $0.61_{\pm 0.01}$ \\
\midrule
\multirow{4}*{NTL-MSE} 
 & 0.15 & $0.09 _{\pm 0.01}$ & $\textbf{57.13}_{\pm 1.51}$ & $\textbf{0.68} _{\pm 0.01}$ \\ 
 &    0.3 & $0.09_{\pm 0.01}$ & $60.99_{\pm 1.35}$ & $0.65_{\pm 0.01}$ \\
 &    0.8 & $0.08_{\pm 0.01}$ & $58.35_{\pm 1.06}$ & $\textbf{0.68}_{\pm 0.01}$ \\
 &    2.0 & $0.07_{\pm 0.01}$ & $59.48_{\pm 2.72}$ & $0.66_{\pm 0.02}$ \\
\midrule
\multirow{4}*{NTL-WAS}
& 0.15 & $0.09 _{\pm 0.00}$ & $57.31 _{\pm 0.57}$ & $\textbf{0.68} _{\pm 0.00}$ \\
&    0.3 & $ \textbf{0.10}_{\pm 0.01}$ & $58.18_{\pm 1.89}$ & $\textbf{0.68}_{\pm 0.02}$ \\
 &    0.8 & $\textbf{0.10}_{\pm 0.02}$ & $59.46_{\pm 1.38}$ & $0.66_{\pm 0.01}$ \\
 &    2.0 & $\textbf{0.10}_{\pm 0.01}$ & $62.29_{\pm 1.04}$ & $0.64_{\pm 0.01}$ \\
 \midrule
NTL-Huber &    0.3 & $\textbf{0.10}_{\pm 0.01}$ & $58.81_{\pm 1.67}$ & $0.67_{\pm 0.01}$ \\
\midrule
NTL-MAE &    0.3 & $\textbf{0.10}_{\pm 0.01}$ & ${57.99}_{\pm 1.48}$ & $0.67_{\pm 0.01}$ \\
\bottomrule
\end{tabular}
    }
\end{minipage}
\label{tab:ablstudies_both}
\end{table}

First, it is evident that all flavors of NTL are consistently better than a standard CE loss.
Mean and standard deviation were calculated over 4 different training runs. Details and additional metrics can be found in Appendix~\ref{Appendix:ablation-studies}.
The ablation studies in~\autoref{tab:ablstudies_both} support our finding that both variants of the NTL improve arithmetic performance compared to using the CE loss alone. Additionally, NTL-WAS mostly outperforms NTL-MSE across both interpolation and extrapolation test sets. Among the tested weights for the NTL, $\lambda=0.3$ yields the best performance for both loss variants. 
Interestingly, we find that NTL-Huber achieves competitive and NTL-MAE sometimes even superior results to NTL-WAS. 
Across all experiments, the R\textsuperscript{2} achieves high values, seemingly in contrast to the accuracy. 
A closer investigation of the error distribution revealed that this can be attributed to many predictions being numerically very close to the ground truth, despite not being strictly accurate. These smaller errors have a very limited impact on R\textsuperscript{2}, while still affecting accuracy.


\subsubsection{Gaussian Cross Entropy}

The Gaussian Cross Entropy (GCE) proposed by~\citet{wang2025tokenmol} constitutes an alternative, yet complimentary means to convey proximity between number tokens.
We argue that it is less principled because it essentially blurs the labels and assigns mass to technically incorrect tokens.
Yet, we assessed its performance by replacing the vanilla CE objective with a GCE objective. We found $\sigma=0.5$ to yield the best results (\autoref{tab:gce_full}).

The experiments on both the interpolation and extrapolation sets suggest that while both NTL and GCE improve model performance, NTL is generally preferable, as it leads to larger improvements.
\begin{table}[ht]
\centering
\caption{
\textbf{Gaussian Cross Entropy (GCE).} 
Standalone and combinatory effect of GCE and NTL. Means and standard deviations across four runs.
}
\vspace{1em}
\tabcolsep=0.1cm
\begin{tabular}{ccccc}
\toprule
\textbf{GCE} & \textbf{NTL} & \textbf{Accuracy} & \textbf{MAE} & \textbf{R2} \\
\midrule
\multicolumn{5}{c}{Interpolation} \\
\midrule
\xmark & \xmark & $0.34$ & $2.15$ & $0.95$ \\
\xmark & \checkmark & $0.43$ & $0.91$ & $\mathbf{0.99}$ \\
\checkmark & \xmark & $0.42$ & $0.95$ & $\mathbf{0.99}$ \\
\checkmark & \checkmark & $\mathbf{0.48}$ & $\mathbf{0.76}$ & $\mathbf{0.99}$ \\
\midrule
\multicolumn{5}{c}{Extrapolation} \\
\midrule
\xmark & \xmark & $0.05$ & $61.92$ & $0.61$ \\
\xmark & \checkmark & $\mathbf{0.10}$ & $\mathbf{58.18}$ & $\mathbf{0.68}$ \\
\checkmark & \xmark & $\mathbf{0.10}$ & $58.55$ & $0.65$ \\
\checkmark & \checkmark & $\mathbf{0.10}$ & $66.67$ & $0.59$ \\
\bottomrule
\end{tabular}
\label{tab:gce}
\end{table}
Furthermore, we investigated the performance of using NTL and GCE in combination. Since the labels are no longer one-hot encodings in this setting, we employed the NTL-WAS-CDF formulation (\autoref{eq:ntl-was-ot}) which allows for training with arbitrary target distributions. On the interpolation set, the combination of NTL and GCE yields the best results. These experiments suggest that NTL and GCE can be mutually beneficial for certain tasks.

\FloatBarrier
\subsection{Training with NTL Is Efficient}
\begin{figure}[!htb]
    \centering
    \includegraphics[width=\columnwidth]{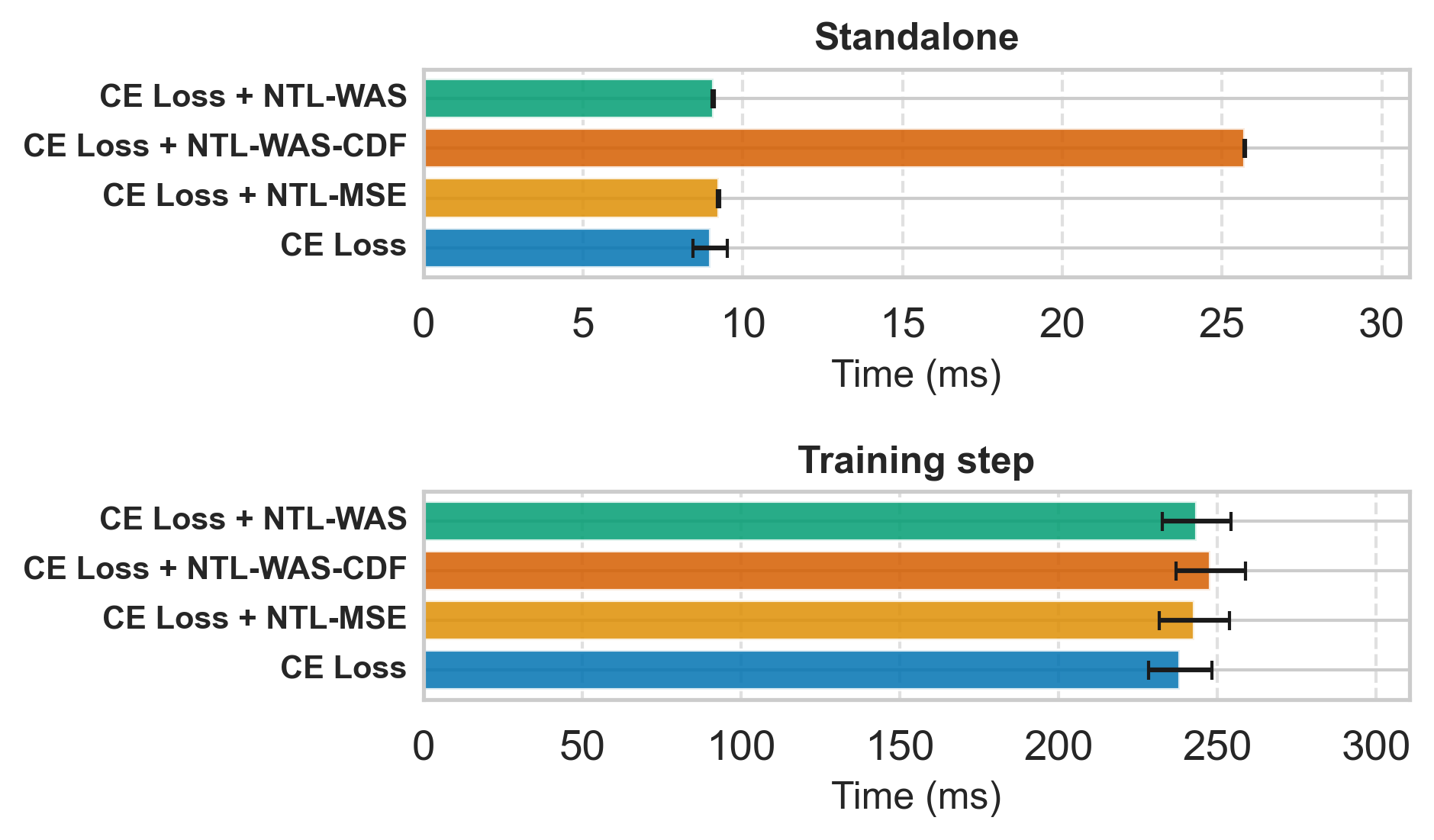}
    \caption{
    \textbf{Runtime Comparison.}
    Benchmarking of the four loss functions in two configurations: standalone execution (top) and a complete training step (bottom), with a mean number token proportion of 80 percent, highlighting computational overheads.}
    \label{fig:Link result}
\end{figure}

To assess potential computational overhead, we performed a comprehensive benchmarking across four distinct loss configurations: \acs{ce} loss serving as the baseline, \acs{ce} with NTL-MSE, \acs{ce} with NTL-Wasserstein implemented according to \autoref{eq:ntl-was-ot} (NTL-WAS-CDF), and \acs{ce} with its optimized variant (NTL-WAS, \autoref{eq:ntl-was}).
We evaluated these loss functions under two different scenarios to quantify both the standalone computational overhead of the loss calculation and its impact within a complete training step, including the forward pass, backpropagation and optimizer step.
We standardized the number of tokens in the input for both forward passes and training steps, along with maintaining a consistent proportion of number tokens across all experiments, although their positions were randomized. 
Each loss function was evaluated through 100 iterations on a GPU. 
The results in \autoref{fig:Link result} demonstrate that NTL-MSE and NTL-WAS introduce only marginal computational overhead across both benchmark scenarios, as each benchmark includes the runtime of the CE loss combined with the runtime of the respective NTL variant. 

Notably, computing NTL-WAS alone is around $125$x faster than computing CE alone.
This is achieved because only a tiny fraction of tokens in the vocab are number tokens.
When assessing only the loss computation time, augmenting the \acs{ce} loss with NTL introduces a computational overhead ranging from a 1\% (NTL-WAS) over 2.9\% (NTL-MSE) and up to a substantial 286\% for the less efficient NTL-WAS-CDF.
However, during the full training step, the relative overhead becomes negligible as the overall computational effort consisting of the forward pass, backpropagation and optimizer step becomes dominant. For instance, the relative overhead of NTL-WAS-CDF compared to the \acs{ce} decreases to $4.4\%$.

\subsection{NTL Can Match Regression Models}
Ideally, LMs should be able to solve even tasks that are focused solely on predicting numerical values, such as estimating a property of a molecule. 
Such tasks are often approached with regression heads on LMs which allow to use appropriate loss functions such as the MSE. 
To test the ability of NTL to solve a regression task solely token-based, we train on the rJokes dataset~\cite{weller2020rjokes} that poses the challenge to predict the level of humor in a joke. A train-test split is provided with the dataset.

In \autoref{tab:reghead}, we show the results of the T5 model with \acs{ce}, NTL-WAS and combined with a regression head, as well as the results from the rJokes benchmark~\cite{weller2020rjokes}.
We evaluated all flavors of the T5 model over three independent runs to ensure robust metrics. NTL significantly improves over standard \acs{ce} and performs on-par with the regression-head model. This is remarkable considering that it operates on a token-level and can still be utilized for non-numeric tasks, unlike models that rely on a dedicated regression head. 
Even though BERT, RoBERTA, and XLNet with a regression head outperform T5 trained with NTL-WAS, the performance gap is not significant considering the difference in the number of parameters. 
We attribute this to the fact that these models have nearly twice as many parameters as our T5-small model. 
Details and additional metrics can be found in Appendix \ref{Appendix:regression+nlp}.

\label{results-jokes-dataset}

\begin{table}[htb!]
\caption{
\textbf{NTL Performance on a Regression Task.}
NTL performs on-par with a regression-head model. 
Mean and standard deviations across five runs. 
BERT, RoBERTA and XLNet values are taken from~\citet{weller2020rjokes}.
}
\centering
\small 
    \vspace{1em}
    \tabcolsep=0.1cm
    \resizebox{1.0\linewidth}{!}{
    \begin{tabular}{@{}lcc|ccc@{}}
        \toprule
        \textbf{Model} & \textbf{Loss}   & \textbf{Reg. Head} & \textbf{RMSE}  & \textbf{Pearson} \\
        \midrule
        T5 & CE & \xmark & $2.01_{\pm 0.01}$ & $0.41_{\pm 0.00}$\\
        T5 & NTL-WAS & \xmark   & $1.81_{\pm 0.01}$ & $0.44_{\pm 0.00}$\\  
        T5 & MSE & \cmark &  $1.82_{\pm 0.01}$ & 
        $0.45_{\pm 0.00}$\\
        \midrule
        BERT & MSE & \cmark & 1.62 & {0.47}\\
        RoBERTA & MSE & \cmark &{1.61} & {0.47}\\
        XLNet & MSE & \cmark & 1.74 & 0.46\\
    \bottomrule
    \end{tabular}
    }
    \label{tab:reghead}
\end{table}

\FloatBarrier
\subsection{NTL Is Model-Agnostic}
\label{SEC:Decoder-only}
As a loss function, NTL can be applied to train arbitrary models, including LMs beyond Transformers like Mamba. The experiments described so far were conducted with the T5 model, which is based on an encoder-decoder Transformer architecture. In this section, we demonstrate the effectiveness of training the decoder-only models GPT-2~\cite{radford2019language} and IBM Granite~\cite{ibm2024granite} with NTL. We test model sizes from 125M to 2B.

To this end, we constructed an arithmetic multiplication task, much akin to the length generalization task~\cite{jelassi2023lengthgeneralizationarithmetictransformers}. This tasks consists of multiplying two numbers with $k$ and $l$ digits, with $k,l\in[1,..,5]$ in training and $k,l\in[1,..,6]$ in testing.  We report the mean absolute percentage error (MAPE) separately for unseen interpolation samples, which involve multiplying numbers with up to $5\times5$ digits, and extrapolation samples, which involve multiplications with at least one six-digit factor.
The results presented in~\autoref{tab:ce_ntl_with_sizes} demonstrate the effectiveness of training decoder-only models with NTL, as it consistently improves performance on the multiplication task in terms of MAPE. 

\begin{table}[!htb]

\begin{tabular}{ll|cc|cc}
\toprule
 &  & \multicolumn{2}{c}{Interpolation} & \multicolumn{2}{c}{Extrapolation} \\
Model & Size & CE & NTL & CE & NTL \\
\midrule
GPT2 Small   & 125M  & 0.55 & \textbf{0.49} & 1.11 & \textbf{1.00} \\
GPT2 Medium  & 350M  & 0.43 & \textbf{0.42} & 0.82 & 0.82 \\
GPT2 Large   & 774M  & 0.39 & \textbf{0.37} & 0.76 & \textbf{0.75} \\
GPT2 XL     & 1.5B  & 0.43 & \textbf{0.40} & 0.83 & \textbf{0.82} \\
Granite 3.2  & 2B    & 0.35 & \textbf{0.21} & 0.60 & \textbf{0.42} \\
Granite 3.1 MOE & 1B  & 0.28 & \textbf{0.15} & 0.68 & \textbf{0.23} \\
\bottomrule
\end{tabular}

\caption{\textbf{NTL Performance for Decoder-only Models.} MAPE for training GPT-2 and IBM Granite variants with CE and NTL on the multiplication task.
All values are error percentages.
}
\label{tab:ce_ntl_with_sizes}
\end{table}

The positive impact of NTL is evident across all model scales. Notably, the improvements from NTL are more pronounced in the extrapolation setting than in interpolation, indicating that training with NTL enhances a model's ability to generalize beyond the training distribution. 
This is particularly relevant due to the poor extrapolation capabilities in mathematical tasks
~\cite{yasaman2022impact}.
Some models, such as Granite 3.1, benefit particularly from training with NTL, showing the largest performance gains among the evaluated models (full evaluation in Appendix~\autoref{fig:multiplication_task}).

\begin{figure}[!htb]
    \centering
    \begin{subfigure}[b]{0.49\linewidth}
    \includegraphics[width=\textwidth]{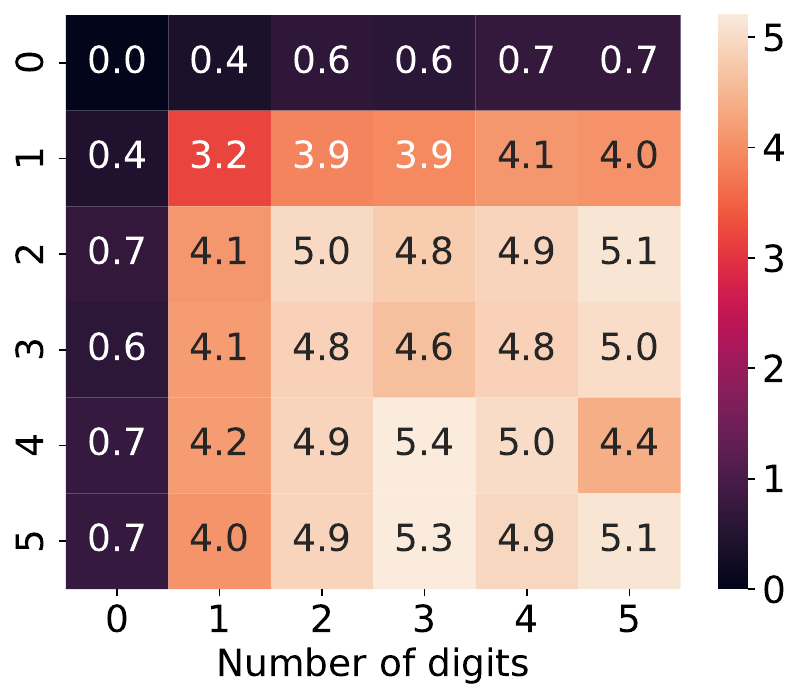}
    \caption{Cross entropy}
    \end{subfigure}
    \begin{subfigure}[b]{0.49\linewidth}
    \includegraphics[width=\textwidth]{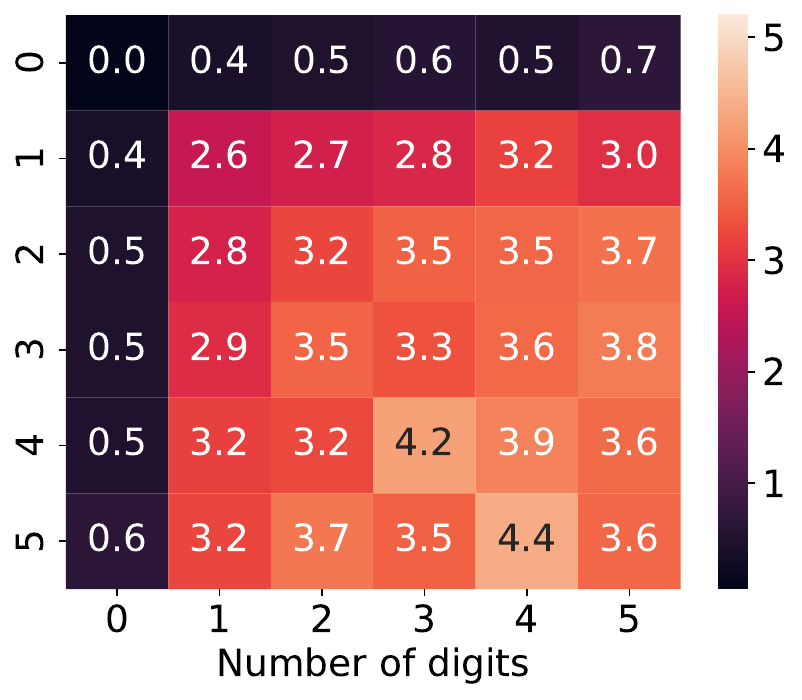}
    \caption{NTL}
    \end{subfigure}
    \caption{\textbf{Sample Efficiency.} Number of epochs needed to reach a MAPE below $0.5$ on the multiplication task (GPT2 Small). NTL particularly improves sample efficiency for samples with longer factors.}
    \label{fig:sample_efficiency}
\end{figure}

Furthermore, we analyzed the sample efficiency of training with NTL on the multiplication task compared to using CE only. As revealed by~\autoref{fig:sample_efficiency}, GPT2 Small trained with NTL requires significantly fewer epochs to achieve a MAPE below $0.5$. This effect is more pronounced for challenging samples consisting of larger factors. 

Similarly, we observe that the average number output distributions are more sharply centered on the correct token when using NTL (see Appendix \autoref{fig:confidence}).
\subsection{NTL Is Effective for Different Tokenizations}
\label{SEC:Tokenizers}

Our standard implementation of NTL-WAS for the T5 model relies on a custom single-digit tokenizer derived from the standard T5 tokenizer, which is based on \texttt{SentencePiece} and includes some tokens consisting of multiple-digits. We conducted additional ablation studies to disentangle the effect of the custom tokenizer and the loss function. To this end, we performed further experiments on the rJokes and the mathematical Q\&A datasets, in which we also evaluated the effect of using the single-digit tokenizer without the NTL-WAS and a more general version of NTL-WAS that supports multi-digit tokens. The results on the rJokes dataset in~\autoref{tab:full_rjokes_evaluation_f} confirm the effectiveness the NTL-WAS loss across both tokenization schemes, with the combination of single-digit tokenization and NTL-WAS performing best.


\begin{table}[h]
\centering
\caption{
\textbf{Tokenizer Comparison on rJokes Datset.}
T5-small used across all experiments. Mean and standard deviation across four runs.
}
\vspace{1em}
\tabcolsep=0.09cm
\label{tab:full_rjokes_evaluation_f}
\begin{tabular}{lccc}
\toprule
\textbf{Loss} & 
\begin{tabular}{@{}c@{}} \textbf{Custom}  \\  \textbf{Tokenizer} \end{tabular}
& \textbf{RMSE} & \textbf{Pearson} \\
\midrule
CE & 
\xmark &
  $2.01_{ \pm 0.01}$ &  
  $0.41_{ \pm 0.00}$ \\
NTL-WAS & 
\xmark &
  $1.97_{ \pm 0.01}$ & 
  $0.41_{ \pm 0.00}$ \\
CE & 
\checkmark &
  $2.02_{ \pm 0.01}$ & 
  $0.41_{ \pm 0.00}$ \\
NTL-WAS & 
\checkmark &
  $\mathbf{1.81_{ \pm 0.01}}$ &  
  $0.44_{ \pm 0.00}$  \\
MSE & 
\xmark &
  $1.82_{ \pm 0.01}$ & 
  $\mathbf{0.45_{\pm 0.00}}$ \\  
\bottomrule
\end{tabular}
\end{table}

A similar trend is observed in the extrapolation test set of the mathematical Q\&A task in~\autoref{tab:tokenizer-ablation-exrapolation}, where the NTL-WAS improves accuracy for both tokenization strategies. 
\begin{table}[htb!]
\centering
\caption{\textbf{Extrapolation Performance for Different Models and Tokenizers}}
\vspace{1em}
\resizebox{\columnwidth}{!}{
\begin{tabular}{l c c c}
\toprule
\textbf{Loss} 
& \begin{tabular}{@{}c@{}} \textbf{Custom} \\ \textbf{Tokenizer} \end{tabular} 
& \begin{tabular}{@{}c@{}}\textbf{Accuracy} \\ \textbf{(Extrapolate)}\end{tabular} 
& \begin{tabular}{@{}c@{}}\textbf{Pearson} \\ \textbf{(Extrapolate)}\end{tabular} \\
\midrule
CE         & \xmark  & $0.05_{\pm 0.00}$ & $0.81_{\pm 0.01}$ \\
NTL-WAS    & \xmark  & $0.06_{\pm 0.00}$ & $0.76_{\pm 0.01}$ \\
CE         & \cmark  & $0.09_{\pm 0.01}$ & $0.87_{\pm 0.01}$ \\
NTL-WAS    & \cmark  & $\mathbf{0.10_{\pm 0.01}}$ & $\mathbf{0.88_{\pm 0.01}}$ \\
\bottomrule
\end{tabular}
}
\label{tab:tokenizer-ablation-exrapolation}
\end{table}
In the interpolation test set in~\autoref{tab:tokenizer-ablation-interpolation}, the NTL-WAS enhances performance for the standard tokenizer, while the single-digit tokenizer performs slightly better without NTL-WAS. Since the extrapolation task and the rJokes task more closely resemble real-world scenarios, we argue that these experiments further reinforce the overall benefits of the NTL, particularly when combined with single-digit tokenization.

Furthermore, we observed a practical issue with models that tokenize numbers into multiple digits: some tokens have large numerical values, which disproportionately affect the loss, especially when the number tokens are irregularly spaced.
For digit-level tokenization and NTL with euclidean distance (as used thus far) the loss for the farthest incorrect number token cannot exceed $9$x of the loss for the nearest incorrect number token\footnote{For a ground truth of \texttt{0}, the nearest/farthest incorrect tokens are \texttt{1} and \texttt{9}.}.
But a single, large number token in the vocabulary, say \texttt{1001}, can distort this ratio to $1000$x and may even destabilize training. 
NTL can intrinsically mitigate such edge cases because it is not limited to using Euclidean distances between numbers (cf.~\autoref{general_wasserstein}).
This can be accomplished through an optional rescaling factor that squashes the distances between tokens.
Note that, with digit-level tokenization, the implicit squashing factor is $9$ while setting it to $1$ effectively recovers standard cross-entropy.
Thus squashing allows to smoothly interpolate between vanilla NTL and CE.
On the integer multiplication dataset~\cite{jelassi2023lengthgeneralizationarithmetictransformers}, we found that GPT2-Large with a squashing factor of $3$ improves over CE but performs worse than NTL as measured by MAPE (\autoref{tab:squashing_results}). 
Overall, we recommend to enforce digit-level tokenization as this ensures NTL is well-behaved.

\subsection{NTL Does Not Hamper Text Learning}
\label{results-multirc-dataset}

To ensure that training with NTL does not introduce drawbacks for normal text generation tasks, we conduct experiments on the MultiRC dataset~\cite{khashabi2018looking}, a question-answering dataset testing reading comprehension. 
We reformat the dataset to train the model to generate the answer based on the question, rather than evaluating it as a multiple-choice task. To this end, we preprocess the dataset by extracting passages along with their corresponding questions and concatenating them to construct the input. 
The answer field consists of all correct answers concatenated with a delimiter. 
The NTL was only applied to the number tokens in the answers, which account for only $1.5\%$ of all answer tokens -- a propotion we believe to be roughly representative of real-world text datasets.
In~\autoref{tab:t5_comparison} the mean and standard deviation over three runs evaluated on the validation set are shown. 
\begin{table}[htb!]
\centering
\caption{
\textbf{NTL on a Standard Language Modeling Task}. 
NTL does not impact performance on a pure textual task.}
\vspace{1em}
\resizebox{\linewidth}{!}{
\begin{tabular}{lccc}
\toprule
\textbf{Loss} & \textbf{Token Acc.} & \textbf{BLEU} & \textbf{ROUGE-1}\\
\midrule
CE
& ${0.36_{\pm 0.00}}$ 
& 0.13$_{\pm 0.00}$
& ${0.32_{\pm 0.00}}$ \\
NTL-WAS
& ${0.36_{\pm 0.00}}$ 
& ${0.13_{\pm 0.00}}$ 
& $0.32_{\pm 0.00}$ \\
\bottomrule
\end{tabular}
}
\label{tab:t5_comparison}
\end{table}
Token accuracy as well as BLEU and ROUGE-1 score remain unaffected by the addition of the NTL and remain on the same level as when only using CE. These results demonstrate that training with NTL does not hamper performance on text comprehension in a task where its benefits for numerical data are limited or absent. 
These findings underline that NTL can be integrated as a standard enhancement in LM training, without compromising general text understanding and generation.

\subsection{NTL Scales Well to LLM-Size}
\label{results-gsm8k-dataset}

To demonstrate the potential to integrate NTL into LLMs, we train the 3B parameter version of the T5 model on the GSM8k dataset~\cite{cobbe2021training}, a benchmark for mathematical reasoning based on school math problems. This dataset poses a significant challenge for standard LLMs; for instance, 6B-parameter models initially achieved only 22\% accuracy after fine-tuning~\cite{cobbe2021training}. In our experiments, the T5-3B model trained on GSM8K with \acs{ce} loss attains a top-1-accuracy of 13.5\% on GSM8K (see \autoref{tab:gsm8k}).
Training with the NTL loss improves accuracy to 17.7\%, matching the performance of LLMs that are more suitable for mathematical reasoning, such as Gemma-2B~\cite{team2024gemma}, and surpassing larger models such as LLaMA-2 7B (14.6\%). A qualitative example of a GSM8K task and predicted solutions is provided in Appendix Example~\autoref{ex:example1}.

\begin{table}[htb!]
\centering
\caption{
\textbf{Results with T5-3B on GSM8k}
}
    \centering
    \vspace{1em}
    \begin{tabular}{cccc@{}}
        \toprule
        \textbf{Loss}    & \textbf{Accuracy}  & \textbf{Pearson} \\
        \midrule
        CE  & $13.5\%$ & $0.67$ \\
        NTL-WAS & $\textbf{17.7}\%$ & $\textbf{0.72}$ \\
    \bottomrule
    \end{tabular}
    \label{tab:gsm8k}
\end{table}

Notably, specialized training approaches can significantly boost performance. For example, tailored fine-tuning strategies have enabled even a 1.3B model to reach 81.5\% accuracy~\cite{liu2023tinygsm}. Future work should explore how NTL can further enhance such specialized models.






\section{Conclusion}

We introduced the \acf{ntl} for Language Models as a novel approach to enhance their intrinsic ability to handle numerical data by considering the numerical proximity between tokens. By augmenting the standard cross entropy loss with NTL, we provide a simple yet effective method that integrates seamlessly into existing architectures without requiring additional computational overhead. Our experiments unambiguously demonstrate the effectiveness of the NTL for encoder-decoder and decoder-only architectures. Experiments on several datasets related to mathematics, arithmetics and reasoning showed significant improvements in numerical reasoning, especially in models without specialized numerical embeddings. 
Furthermore, our results on regression datasets revealed that training with NTL allows a LM with a token-head to compete with models designed specifically for regression tasks. 
We also verified that NTL does not compromise capabilities on standard text-only tasks like reading comprehension.
Finally, the scalability of NTL was demonstrated by successfully applying it to a 3B-parameter model on the challenging GSM8k dataset, opening the door for integration into LLMs. 

From the two proposed NTL flavors (NTL-WAS and NTL-$\mathcal{L}_p$) the NTL-WAS is a preferable loss since it avoids the undesired local minima of NTL-$\mathcal{L}_p$ caused by errors canceling out.
However, unlike in NTL-WAS, the dot product inside NTL-$\mathcal{L}_p$ yields a quantity in the original number space which can be useful to compute more complex losses incorporating multiple tokens.
For example, a current limitation of NTL is to treat each digit-level token as equally important, thereby ignoring the higher relevance of digits with higher decimal places. 
This limitation could be addressed by computing the loss at the number level rather than at the digit level, e.g., by naive scaling by decimal place or by leveraging the dot-product idea of NTL-$\mathcal{L}_p$ to calculate a number-level prediction.
Furthermore, future research should focus on exploring synergies with additional arithmetic-specific training adaptations. Particularly exciting will be the fine-tuning of mid- and large-scale open-source models, such as LLaMA or Gemma, to further assess its scalability and generalization potential. 
Overall, this approach offers a practical solution for enhancing language models in numerically rich domains, contributing towards more accurate and reliable applications of LMs in mathematics and science.



\section*{Further Resources}
\subsection*{Code}
\begin{enumerate}[leftmargin=*, topsep=0pt, partopsep=0pt, itemsep=4pt, parsep=0pt]
    \item \textbf{Usage:} NTL ships via PyPI. For minimalistic integration in your work use {\urlstyle{same} \small \href{https://ibm.biz/ntl-pypi-repo}{github.com/AI4SD/number-token-loss}}
    \item \textbf{Reproduction:} Full development code for reproduction of all experiments: 
    {\urlstyle{same} \small \href{https://ibm.biz/ntl-code}{github.com/TUM-AI/number-token-loss}}

\end{enumerate}

\subsection*{Learn more}
\begin{enumerate}[leftmargin=*, topsep=0pt, partopsep=0pt, itemsep=4pt, parsep=0pt]

    \item \textbf{Landing page:} 
    {\urlstyle{same}  \href{https://ibm.biz/ntl-main}{Project overview site}}
    
    \item \textbf{Demo:} {\urlstyle{same} \href{https://ibm.biz/ntl-demo}{Streamlit app on HuggingFace spaces}}

    \item \textbf{Video:} 
    {\urlstyle{same} \href{https://ibm.biz/ntl-5min-yt}{5min paper walkthrough on YouTube}}

\end{enumerate}

\section*{Impact Statement}

The introduction of the NTL enhances the ability of LMs to handle numerical data and can improve performance in tasks involving quantitative reasoning and arithmetic. This advancement has potential implications across fields like scientific research, finance, and engineering, where precise numerical outputs are essential for data-driven decision-making and automated analysis. 
However, this capability also raises potential risks, including misuse of generated numerical data or biased predictions in sensitive areas, such as healthcare or economics. As with any AI methods, careful implementation and human oversight are necessary to ensure model transparency and accuracy.

\newpage
\bibliography{main}
\bibliographystyle{icml2025}
\newpage
\appendix
\onecolumn
\section{Appendix}

\subsection{Statement on code}
The code for this paper is available at \href{https://github.com/tum-ai/number-token-loss}{https://github.com/tum-ai/number-token-loss}.

\subsection{Algorithm for the Number Token Loss}
\label{algo:ntl}
\renewcommand{\thefigure}{A\arabic{figure}}
\renewcommand{\thealgorithm}{A\arabic{algorithm}}
\renewcommand{\thetable}{A\arabic{table}}
\setcounter{figure}{0} 
\setcounter{table}{0} 

\begin{algorithm*}
\caption{Pseudo-code to compute NTL-MSE}
\begin{algorithmic}[1]
\State \textbf{Initialize:} $\text{n\_vocab} 
\leftarrow \left[ \begin{cases}
    \text{int}(\text{vocab}[i]) & \text{if } \text{vocab}[i] \in \mathbb{R} \\
    \text{NaN} & \text{otherwise}
    \end{cases}
    \right]_{i=1}^V$
    
\State 
\Function{Forward}{$\text{logits} \in \mathbb{R}^{B \times T \times V}$, $\text{labels} \in \mathbb{R}^{B \times T}$} : \text{Float}
    \State $\text{ntl} \leftarrow 0$
    \State $\text{n\_logits} \leftarrow \text{logits}[:, :, \neg\text{n\_vocab.isnan()}]$ \Comment{Ignore non-number tokens}
    \State $\text{n\_probs} \leftarrow \text{Softmax}(\text{logits})$
    \State $\hat{y} \leftarrow \sum_{i} \text{n\_probs}[:, :, i] \cdot \text{n\_vocab}$ \Comment{$\hat{y}$ is $B \times T$}
    \State $y \leftarrow \text{n\_vocab}[\text{labels}]$ \Comment{$y$ is $B \times T$}
    \State $\text{ntl} \leftarrow \text{MSE}(y, \hat{y})$ \Comment{Can be any regression loss}
    \State \Return $\text{ntl}$
\EndFunction
\end{algorithmic}
\end{algorithm*}

\begin{algorithm*}
\caption{Pseudo-code to compute NTL-WAS}
\begin{algorithmic}[1]
\State \textbf{Initialize:} $\text{n\_vocab} 
\leftarrow \left[ \begin{cases}
    \text{int}(\text{vocab}[i]) & \text{if } \text{vocab}[i] \in \mathbb{R} \\
    \text{NaN} & \text{otherwise}
    \end{cases}
    \right]_{i=1}^V$
\If{order\_numbers is True}
    \State Sort the numbers in $\text{n\_vocab}$ by their numerical values
\EndIf

\State
\Function{Forward}{$\text{logits} \in \mathbb{R}^{B \times T \times V}$, $\text{labels} \in \mathbb{N}^{B \times T}$} : \text{Float}
     \State $\text{n\_logits} \leftarrow \text{logits}[:, :, \neg\text{n\_vocab.isnan()}]$ \Comment{Ignore non-number tokens}
    \State $\text{n\_probs} \leftarrow \text{Softmax}(\text{logits})$
    \State $y \leftarrow \text{n\_vocab}[\text{labels}]$ \Comment{Retrieve true numerical values}

    \If{using CDF version of NTL-WAS}
    \State $\text{y\_distr}[b, t] \leftarrow \text{one\_hot}(y[b, t], \text{num\_classes=len(n\_vocab)})$ \Comment{One hot encode y}
    \State $\text{wasserstein\_distance}[b, t] = \sum_{v=1}^{V} \left| \text{CDF}(\text{n\_probs}[b, t])[v] - \text{CDF}(\text{y\_distr}[b, t])[v]\right|$
    \EndIf
    \If{using absolute difference version of NTL-WAS}
    \State $\text{wasserstein\_distance}[b, t] = \leftarrow \frac{1}{V} \sum_v^V \text{n\_probs}[b, t, v] \cdot |\text{n\_vocab}[v] - y|$ 
    \EndIf
    \State $\text{ntl} \leftarrow \text{Mean}(\text{wasserstein\_distance}[\neg y.\text{isnan()}])$
    \State \Return $\text{ntl}$
\EndFunction
\end{algorithmic}
\end{algorithm*}

\FloatBarrier
\subsection{T5 Architecture} 
\label{Appendix:T5}
The T5 model is built upon the Transformer architecture \cite{vaswani2017attention}, consisting of stacked self-attention and feed-forward layers in both the encoder and decoder. The encoder processes the input tokens to create contextualized representations, while the decoder generates the output tokens autoregressively, attending to both the encoder's outputs and the previously generated tokens. The model can be trained with both Masked Language Modelling (MLM)~\cite{kenton2019bert} and Causal/Auto-Regressive Language Modelling (CLM)~\cite{dai2015semi}, whereby we chose to use CLM.

\subsection{Regression Transformer}
\label{Appendix:RT}
The Regression Transformer~\cite{born2023regression} preserves the inherent structure of numbers by inducing information on relative proximity through numerical encodings that are set deterministically for all tokens. For every combination of a decimal place and digit value, a corresponding numerical token is added to the vocabulary. For instance, the number $11.4$ is tokenized to \texttt{[1\_1, 1\_0, 4\_-1]}. 

Non-numeric tokens are set to zero vectors. The numerical encodings are designed so that their pairwise distances are symmetric and monotonically decreasing with the float value. The final encoding of the input tokens is obtained by summing over numerical and regular word encodings.
The Regression Transformer numerical encodings NE at dimension $j$ for numerical token $t_{v,p}$ with value $v$ and decimal place $p$ can be determined by
\begin{equation}
    \text{NE}_{\text{Float}}(v, p, j) = (-1)^j \cdot \frac{v \cdot 10^p}{j + 1}.
\end{equation}

\subsection{Challenges with Integrating xVal in Transformer Models like T5}
\label{Appendix:xVal}
In transformer models like T5, integrating numerical encoding schemes like xVal presents challenges. 
xVal multiplyies the \texttt{[NUM]} token embedding $X$ by the number value 
$a$. In T5, however, a per-sample pre-layer normalization is applied immediately after the embedding, which effectively removes the scaling by $a$. Specifically:

\begin{equation}
    \frac{aX-E[aX]}{\sigma(aX)}=\frac{aX-aE(X)}{\sqrt{a^2E(X^2)-a^2E(X)^2}}=\frac{X-E(X)}{\sigma(X)}
\end{equation}
Hence, under T5’s architecture, all numbers collapse to the same embedding, making xVal incompatible with T5.

Even in the original xVal architecture, the range of values xVal can process meaningfully is limited by the layer normalization that follows the positional embedding step. Please see \citet{golkar2023xval} under "Implicit normalization via layer-norm" for more information. Therefore, xVal normalizes each value to $[-5,5]$ prior to training to mitigate this issue.

We argue that this approach cannot be applied in practice, since in real texts the range of numbers is not known in advance, and thus a simple min-max normalisation to $[-5,5]$ prior to training or inference is not really practical.

Therefore we opted for a simpler approach in our experiments: Applying a signed $log(1+x)$ transformation to all numeric inputs. This avoids the overhead of parsing and re-scaling each number to $[-5,5]$ prior to training, but it also has the drawback that large numbers are squashed in the embedding space, making fine-grained distinctions difficult for the model.

For a direct comparison without any modifications to the xVal processing, we repeated the 3-digit multiplication experiment from \citet{golkar2023xval}. The results can be seen in \autoref{tab:r2-encodings}.

\begin{table*}[!htb]
    \centering
    \vspace{1em}
    \begin{tabular}{lc}
        \toprule
        \textbf{Encoding} & \textbf{R\textsuperscript{2} Value} \\
        \midrule
        P10                 & 0.9989   \\ 
        P1000               & 0.9997   \\ 
        B1999               & 0.9998   \\ 
        FP15                & 0.7119   \\ 
        xVal                & 0.9986   \\ 
        T5 CE               & 0.999934 \\ 
        T5 NTL-WAS          & \textbf{0.999997} \\ 
        T5 regression head  & 0.999891 \\ 
        \bottomrule
    \end{tabular}
    \caption{R\textsuperscript{2} scores for various number encoding methods on the 3-digit multiplication experiment from \citet{golkar2023xval}.}
    \label{tab:r2-encodings}
\end{table*}

\FloatBarrier
\subsection{Experiments}\label{app:hyperparameters}
For all trainings, we use \texttt{transformers}~\cite{wolf2020transformers} 4.42.4.
We train with a batch size of 32, a learning rate of 1e-4 and a weight decay of 0.01.
All models were trained on single graphics processing units (GPUs) of type NVIDIA RTX A6000, NVIDIA A100 or NVIDIA A40.

\subsubsection{Multitask Mathematics Dataset}
\label{dataset-details}
To test the mathematical capabilities of the methods, we use a subset of the mathematical question-answer dataset from DeepMind~\cite{saxton2019analysingmathematicalreasoningabilities}.
The dataset was generated synthetically and therefore contains limited linguistic variability, but is sufficient for our purposes to compare the mathematical capabilities of the different methods.

The dataset contains different modules and difficulty levels.
For training and testing the models, we chose all difficulty levels but excluded modules where the answer contains complex fractions or variables. This allows us to focus on purely numeric answers to simplify the evaluation of the model and still leaves us with a large enough dataset of $\sim$26 million samples.

For training, validation, and interpolation tests, we selected the following modules from the DeepMind mathematical question-answer dataset:

\begin{itemize}[itemsep=-1pt]
    \item \texttt{algebra\_\_linear\_1d.txt} 
    \item \texttt{algebra\_\_linear\_1d\_composed.txt} 
    \item \texttt{algebra\_\_linear\_2d.txt} 
    \item \texttt{algebra\_\_linear\_2d\_composed.txt} 
    \item \texttt{algebra\_\_sequence\_next\_term.txt} 
    \item \texttt{arithmetic\_\_add\_or\_sub.txt} 
    \item \texttt{arithmetic\_\_add\_sub\_multiple.txt} 
    \item \texttt{arithmetic\_\_mul.txt} 
    \item \texttt{numbers\_\_div\_remainder.txt} 
    \item \texttt{numbers\_\_div\_remainder\_composed.txt} 
    \item \texttt{numbers\_\_place\_value.txt} 
    \item \texttt{numbers\_\_round\_number.txt} 
    \item \texttt{numbers\_\_round\_number\_composed.txt} 
\end{itemize}

For extrapolation tests, we selected the following modules:

\begin{itemize}[itemsep=-1pt]
    \item \texttt{arithmetic\_\_add\_or\_sub\_big.txt} 
    \item \texttt{arithmetic\_\_add\_sub\_multiple\_longer.txt} 
    \item \texttt{arithmetic\_\_mixed\_longer.txt} 
    \item \texttt{arithmetic\_\_mul\_big.txt} 
    \item \texttt{arithmetic\_\_mul\_div\_multiple\_longer.txt} 
    \item \texttt{numbers\_\_place\_value\_big.txt} 
    \item \texttt{numbers\_\_round\_number\_big.txt} 
\end{itemize}

This resulted in a training dataset of 25,986,948 samples, a validation dataset of 13,026 samples, an interpolation test set of 130,000 samples, and an extrapolation test set of 70,000 samples.

\begin{figure}[!htb]
    \centering
    \begin{subfigure}[b]{0.48\linewidth}
        \centering
        \includegraphics[width=\columnwidth]{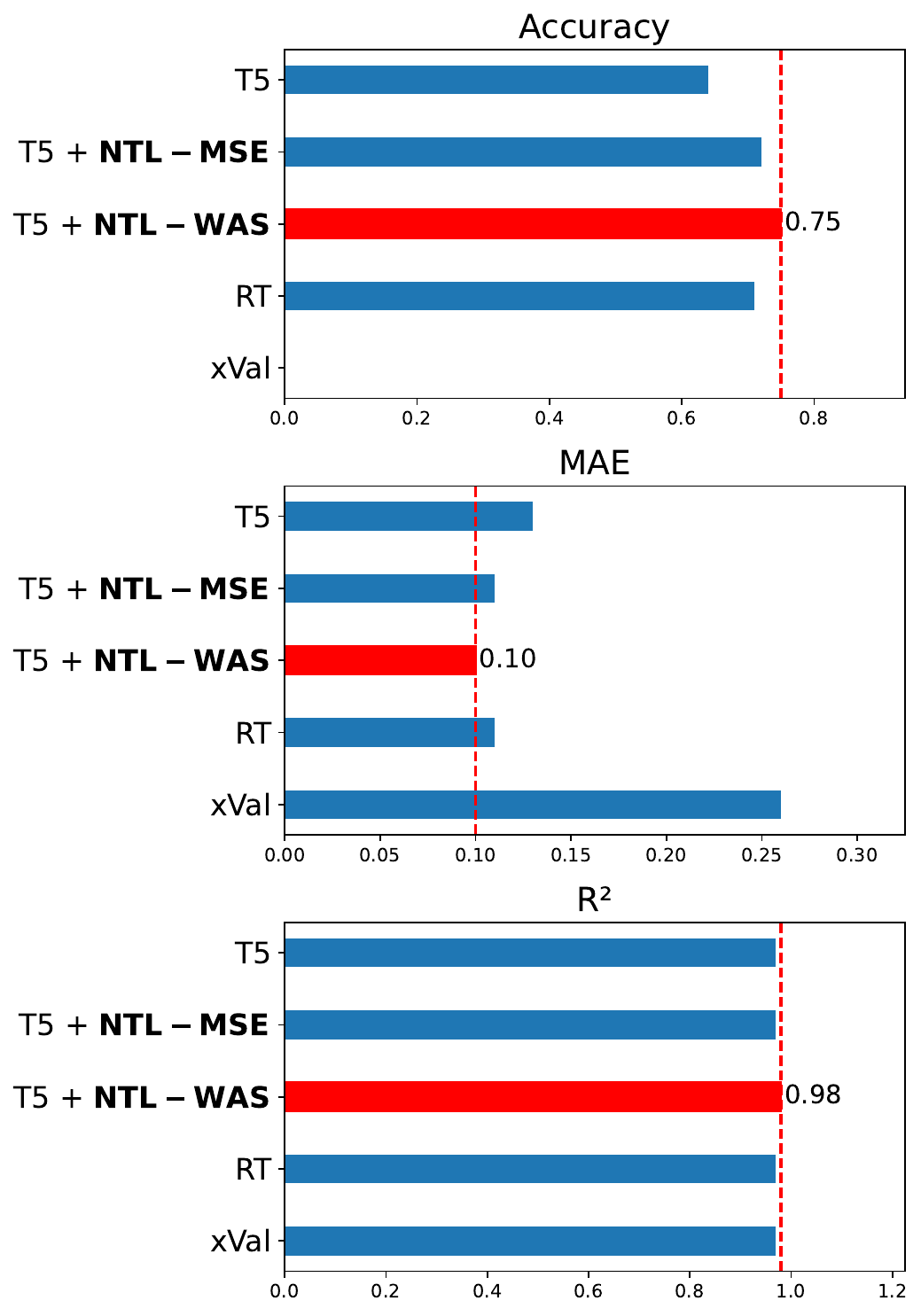}
        \caption{
        Evaluation metrics on interpolated test data.
        }
        \label{fig:res-bar-interpolate}
    \end{subfigure}
    \hfill
    \begin{subfigure}[b]{0.48\linewidth}
        \centering
        \includegraphics[width=\columnwidth]{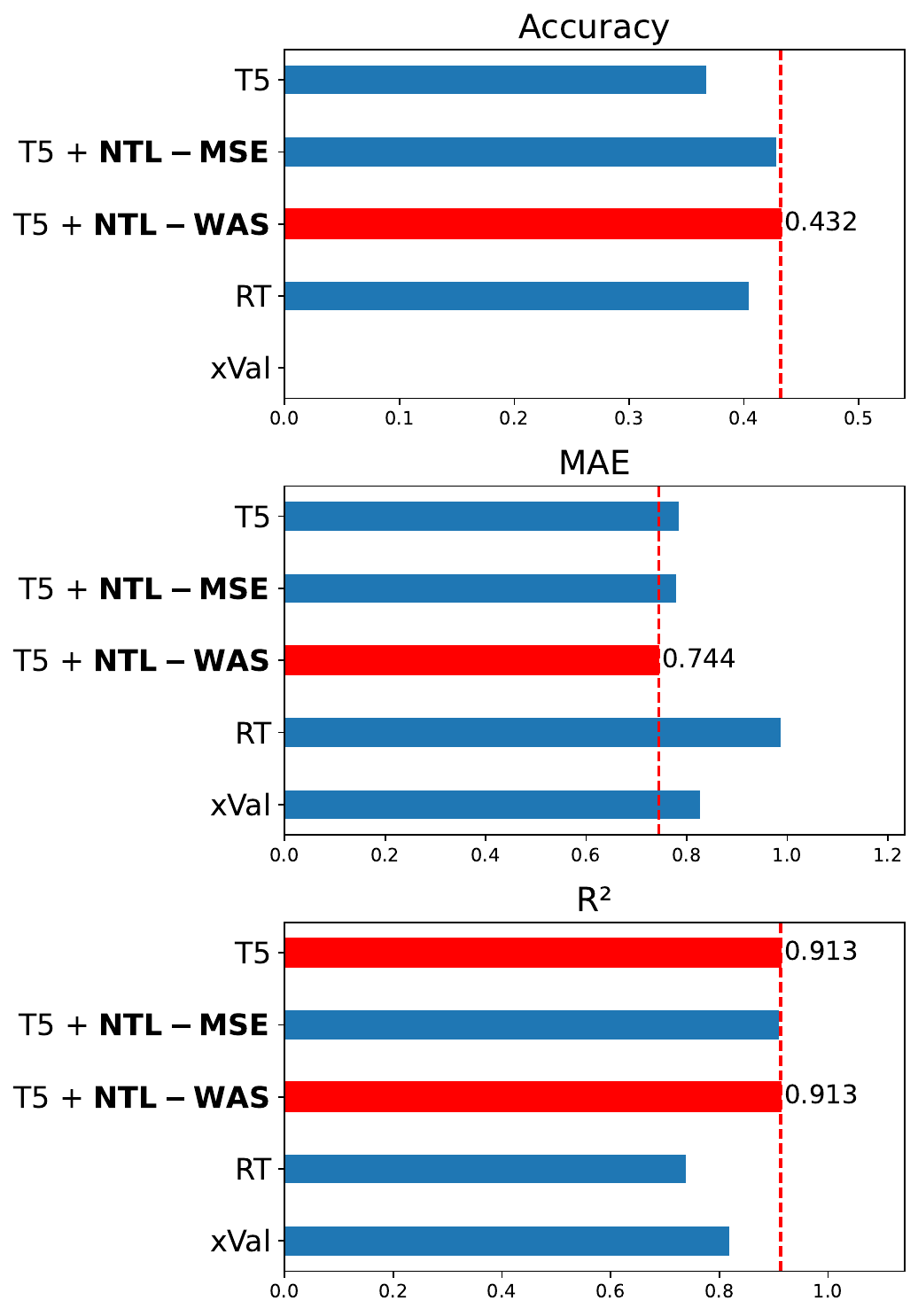}
        \caption{
        Evaluation metrics on extrapolated test data.
        }
        \label{fig:res-bar-extrapolate}
    \end{subfigure}
    \caption{Comparison of evaluation metrics on interpolated and extrapolated test data.
    }
    \label{fig:res-bar-comparison}
\end{figure}

We train each model for 1050000 iterations. For these experiments, we used the T5-base architecture (220M parameters).
For the Number Token Loss, we trained with the hyperparameter $\lambda$ set to 0.3.
The results can be seen in~\autoref{tab:table_result} and~\autoref{fig:res-bar-comparison}.

\subsubsection{Ablation studies and NTL-WAS-CDF with Gaussian smoothing}
\label{Appendix:ablation-studies}
For training and validation, we selected a subset of the DeepMind mathematical Q\&A dataset, in all three difficulty levels: \texttt{arithmetic\_\_add\_sub\_multiple.txt}.
Similarly, interpolation and extrapolation tests were done on \texttt{arithmetic\_\_add\_sub\_multiple.txt} and \texttt{arithmetic\_\_add\_sub\_multiple\_longer.txt}, respectively. 
This resulted in a training set of 100,000 examples, a validation set of 3,000 examples, and two test sets of 10,000 examples each.

For these experiments, we used the T5-small architecture (60M parameters). The mean and standard deviation were taken over 4 different runs for the ablation studies and 3 different runs for the studies on GCE.
We trained each run for 2.5M iterations. 

In addition to accuracy, mean absolute error and R\textsuperscript{2} given in ~\autoref{tab:ablstudies_interpolate} and \autoref{tab:gce}, the achieved Pearson correlation and Spearman's rank correlation coefficient are shown in ~\autoref{tab:abl-studies-pear-spear}.

\begin{table*}[htb!]
\centering

\caption{
\textbf{Pearson and Spearman Correlation Coefficients on the Interpolation and Extrapolation Test Sets}
}

    \centering
    \vspace{1em}
    \begin{tabular}{lccccc}
        \toprule
        \textbf{Loss} & $\lambda$ & \multicolumn{2}{c}{\textbf{Pearson}} & \multicolumn{2}{c}{\textbf{Spearman}}  \\
        &  & Interpolation & Extrapolation & Interpolation & Extrapolation \\
        \midrule
    
        CE & & $0.98 _{\pm 0.00}$ & $0.81 _{\pm 0.01}$ & $0.98 _{\pm 0.00}$ & $0.86 _{\pm 0.01}$ \\ 
        \midrule
        \multirow{3}*{NTL-MSE} & 0.3 & $1.00 _{\pm 0.00}$ & $0.86 _{\pm 0.00}$ & $1.00 _{\pm 0.00}$ & $0.89 _{\pm 0.00}$ \\ 
        & 0.8 & $0.99 _{\pm 0.00}$ & $0.87 _{\pm 0.00}$ & $1.00 _{\pm 0.00}$ & $0.90 _{\pm 0.00}$ \\ 
        & 2.0 & $0.99 _{\pm 0.00}$ & $0.86 _{\pm 0.01}$ & $0.99 _{\pm 0.00}$ & $0.89 _{\pm 0.01}$ \\ 
        \midrule
        
        \multirow{3}*{NTL-WAS} & 0.3 & $1.00 _{\pm 0.00}$ & $0.88 _{\pm 0.01}$ & $1.00 _{\pm 0.00}$ & $0.91 _{\pm 0.01}$ \\ 
        & 0.8 & $1.00 _{\pm 0.00}$ & $0.86 _{\pm 0.00}$ & $1.00 _{\pm 0.00}$ & $0.89 _{\pm 0.00}$ \\ 
        & 2.0 & $0.99 _{\pm 0.00}$ & $0.85 _{\pm 0.01}$ & $1.00 _{\pm 0.00}$ & $0.88 _{\pm 0.01}$ \\ 
        \midrule
        NTL-MAE & 0.3 & $1.00 _{\pm 0.00}$ & $0.87 _{\pm 0.01}$ & $1.00 _{\pm 0.00}$ & $0.90 _{\pm 0.01}$ \\ 
        \midrule
        NTL-Huber & 0.3 & $1.00 _{\pm 0.00}$ & $0.87 _{\pm 0.01}$ & $1.00 _{\pm 0.00}$ & $0.90 _{\pm 0.01}$ \\ 
        \midrule
        {GS + CE} & & $0.49 _{\pm 0.38}$ & $0.20 _{\pm 0.22}$ & $1.00 _{\pm 0.00}$ & $0.90 _{\pm 0.01}$ \\ 
  
        {GS + NTL-WAS} & 0.3 & $1.00 _{\pm 0.00}$ & $0.87 _{\pm 0.01}$ & $1.00 _{\pm 0.00}$ & $0.90 _{\pm 0.01}$ \\ 

        \bottomrule
    \end{tabular}
    \label{tab:abl-studies-pear-spear}
\end{table*}

\subsubsection{NTL for regression and normal NLP task}
\label{Appendix:regression+nlp}
We trained and evaluated the NTL on two tasks: a regression task using the rJokes dataset~\cite{weller2020rjokes} and a test generation task using the transformed MultiRC dataset~\cite{khashabi2018looking}. In both cases, each model was trained for 2.5M iterations. The mean and standard deviation were taken over 3 different runs.
When using T5 with NTL-WAS, we set the hyperparameter $\lambda$ to 2.0.

In addition to RMSE and Pearson correlation given in \autoref{tab:full_rjokes_evaluation_f}, mean squared error, accuracy and Spearman’s rank correlation coefficient are shown in \autoref{tab:full_rjokes_evaluation}.

\begin{table*}[htb]
\centering
\caption{
\textbf{Full Evaluation on the rJokes Dataset.} Evaluation metrics (mean $\pm$ std).
}
\label{tab:full_rjokes_evaluation}
\vspace{1em}
\begin{tabular}{@{}lcccccc@{}}
\toprule
\textbf{Model} & \textbf{MSE} & \textbf{Number accuracy} & \textbf{Pearson} & \textbf{Spearman} \\
\midrule
{Standard T5 (CE)} & 
  4.03 \footnotesize{ $\pm$ 0.03} &  
  0.30 \footnotesize{ $\pm$ 0.00} & 
  0.41 \footnotesize{ $\pm$ 0.00} & 
  0.35 \footnotesize{ $\pm$ 0.00} \\
{T5 + NTL} & 
  $\mathbf{3.27 \footnotesize{ \pm 0.04}}$ &  
  0.30 \footnotesize{ $\pm$ 0.00} & 
  0.44 \footnotesize{ $\pm$ 0.00} & 
  $\mathbf{0.40}$ \footnotesize{ $\mathbf{\pm 0.00}$} \\
{T5 + Regression Head (MSE)} & 
  3.32 \footnotesize{ $\pm$ 0.04} & 
  0.00 \footnotesize{ $\pm$ 0.00} & 
  $\mathbf{0.45}$ \footnotesize{ $\mathbf{\pm 0.00}$} & 
  $\mathbf{0.40}$ \footnotesize{ $\mathbf{\pm 0.00}$} \\  
{T5 + NTL + Standard tokenizer} & 
  3.88 \footnotesize{ $\pm$ 0.02} & 
  0.30 \footnotesize{ $\pm$ 0.00} & 
  0.41 \footnotesize{ $\pm$ 0.00} & 
  0.35 \footnotesize{ $\pm$ 0.00} \\
{Standard T5 (CE) + Custom Tokenizer} & 
  4.09 \footnotesize{ $\pm$ 0.04} & 
  0.30 \footnotesize{ $\pm$ 0.00} & 
  0.41 \footnotesize{ $\pm$ 0.00} & 
  0.37 \footnotesize{ $\pm$ 0.00} \\
\bottomrule
\end{tabular}
\end{table*}

\subsubsection{ NTL is effective for different tokenizations}
To evaluate the effect of the tokenizer we trained T5-small again for 2.5M iterations.
For the NTL-WAS loss we set the hyperparameter $\lambda$ to 2.0. A comparison of the achieved number accuracies and Pearson correlation of the models is shown in~\autoref{tab:tokenizer-ablation-interpolation},~\autoref{tab:tokenizer-ablation-exrapolation} and~\autoref{tab:full_rjokes_evaluation}.

\begin{figure}[htb!]
    \centering
    \includegraphics[width=0.4\linewidth]{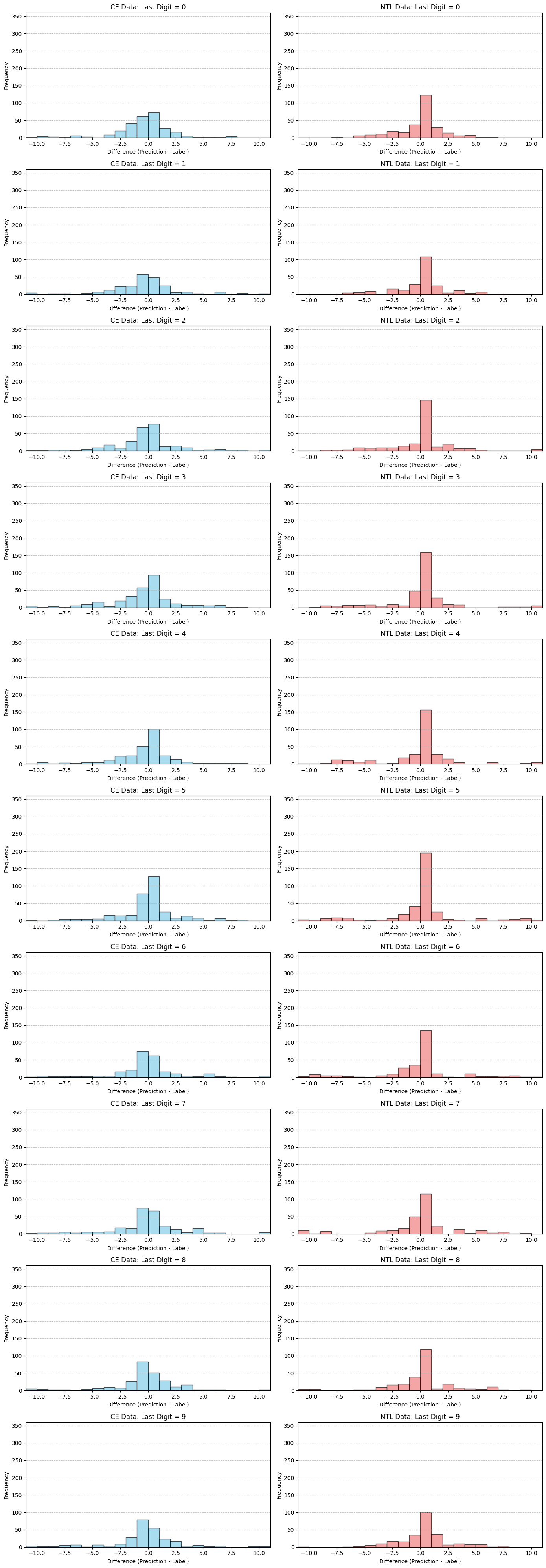}
    \caption{Error histograms for all number tokens for the GSM8k task.}
    \label{fig:histograms}
\end{figure}

\subsubsection{Error Analyis}

we conducted a detailed error analysis on the GSM8K dataset to examine predictions for numbers ending with specific digits (0–9), comparing CE and NTL. The error histograms shown in~\autoref{fig:histograms} reveal a consistent pattern across all digit groups: NTL error distributions are narrower and concentrated around zero, confirming improved numerical reasoning compared to CE.

We further analyzed model predictions for numbers near digit boundaries, focusing particularly on numbers ending with $0$ or $9$ as well as powers of 10. \autoref{tab:error_rates} reveals how often predictions are overestimations, underestimations and exact matches. The results show, that NTL achieves a more balanced error distribution. The exact match accuracy is consistently higher, but particularly so for samples ending with the 9 token, implying that NTL leads to a better handling on those digit boundaries.

\begin{table}[h]
\centering
\caption{\textbf{Error Rates.} Error Rates by Digit Boundary Sample Type on GSM8K.}
\label{tab:digit-boundary-errors}
\begin{tabular}{l l c c}
\toprule
\textbf{Sample Type} & \textbf{Metric} & \textbf{CE (\%)} & \textbf{NTL (\%)} \\
\midrule
\multirow{3}{*}{Ends with 0} 
  & Overestimation Rate     & $\mathbf{28.4}$ & $29.4$ \\
  & Underestimation Rate    & $56.5$ & $\mathbf{51.1}$ \\
  & Exact Match Rate        & $15.0$ & $\mathbf{19.4}$ \\
\midrule
\multirow{3}{*}{Power of 10} 
  & Overestimation Rate     & $54.4$ & $\mathbf{49.1}$ \\
  & Underestimation Rate    & $24.6$ & $24.6$ \\
  & Exact Match Rate        & $21.1$ & $\mathbf{26.3}$ \\
\midrule
\multirow{3}{*}{Ends with 9} 
  & Overestimation Rate     & $\mathbf{33.3}$ & $38.8$ \\
  & Underestimation Rate    & $46.7$ & $\mathbf{32.6}$ \\
  & Exact Match Rate        & $20.0$ & $\mathbf{28.6}$ \\
\bottomrule
\label{tab:error_rates}
\end{tabular}
\end{table}

\begin{table}[htb!]
\centering
\caption{\textbf{Interpolation Performance for Different Models and Tokenizers}}
\vspace{1em}
\begin{tabular}{l c c c}
\toprule
\textbf{Loss} 
& \begin{tabular}{@{}c@{}} \textbf{Custom} \\ \textbf{Tokenizer} \end{tabular} 
& \begin{tabular}{@{}c@{}}\textbf{Accuracy} \\ \textbf{(Interpolate)}\end{tabular} 
& \begin{tabular}{@{}c@{}}\textbf{Pearson} \\ \textbf{(Interpolate)}\end{tabular} \\
\midrule
CE         & \xmark  & $0.34_{\pm 0.01}$ & $0.98_{\pm 0.00}$ \\
NTL-WAS    & \xmark  & $0.39_{\pm 0.00}$ & $0.96_{\pm 0.00}$ \\
CE         & \cmark  & $\mathbf{0.45_{\pm 0.01}}$ & $\mathbf{1.00_{\pm 0.00}}$ \\
NTL-WAS    & \cmark  & $0.43_{\pm 0.05}$ & $\mathbf{1.00_{\pm 0.00}}$ \\
\bottomrule
\end{tabular}
\label{tab:tokenizer-ablation-interpolation}
\end{table}

\begin{table}[ht]
\centering
\caption{
\textbf{Gaussian Cross Entropy (GCE).} 
Standalone and combinatory effect of GCE and NTL.
}
\vspace{1em}
\tabcolsep=0.1cm
\begin{tabular}{cccccc}
\toprule
\textbf{GCE} &$\boldsymbol{\sigma}$&\textbf{NTL} & \textbf{Accuracy} & \textbf{MAE} & \textbf{R2} \\
\midrule
\multicolumn{6}{c}{Interpolation test set} \\
\midrule
\xmark & - &\xmark & $0.34$ & $2.15$ & $0.95$ \\
\xmark & - & \checkmark & $0.43$ & $0.91$ & $\mathbf{0.99}$ \\
\checkmark & 1 &\xmark & $0.19$ & $4.34$ & $0.95$ \\
\checkmark & 1 &\checkmark & $0.34$ & $3.14$ & $0.96$ \\
\checkmark & 0.5 &\xmark & $0.42$ & $0.95$ & $\textbf{0.99}$ \\
\checkmark & 0.5 &\checkmark & $\textbf{0.48}$ & $\textbf{0.76}$ & $\textbf{0.99}$ \\
\midrule
\multicolumn{6}{c}{Extrapolation test set} \\
\midrule
\xmark & - &\xmark & $0.05$ & $61.92$ & $0.61$ \\
\xmark & - & \checkmark & $\textbf{0.10}$ & $\textbf{58.18}$& $\textbf{0.68}$ \\
\checkmark & 1 & \xmark & $0.03$ & $126.74$ & $0.08$ \\
\checkmark & 1 & \checkmark & $0.06$ & $111.09$ & $0.25$ \\
\checkmark & 0.5 & \xmark & $\textbf{0.10}$ & $58.55$ & $0.65$ \\
\checkmark & 0.5 & \checkmark & $\textbf{0.10}$ & $66.97$ & $0.59$ \\
\bottomrule
\end{tabular}
\label{tab:gce_full}
\end{table}

\begin{table}[htb!]
\centering
\begin{tabular}{lc}
\toprule
\textbf{Loss} & \textbf{MAPE} \\
\midrule
CE             & $0.502$\% \\
NTL-Squash-2   & $0.491$\% \\
NTL            & $\mathbf{0.485}$\% \\
\bottomrule
\end{tabular}
\caption{\textbf{Squashing NTL Loss.} MAPE on the multiplication task using the GPT-2 model trained with a squashing NTL variant.}
\label{tab:squashing_results}
\end{table}

\begin{figure}[htb!]
    \centering
    \includegraphics[width=\linewidth]{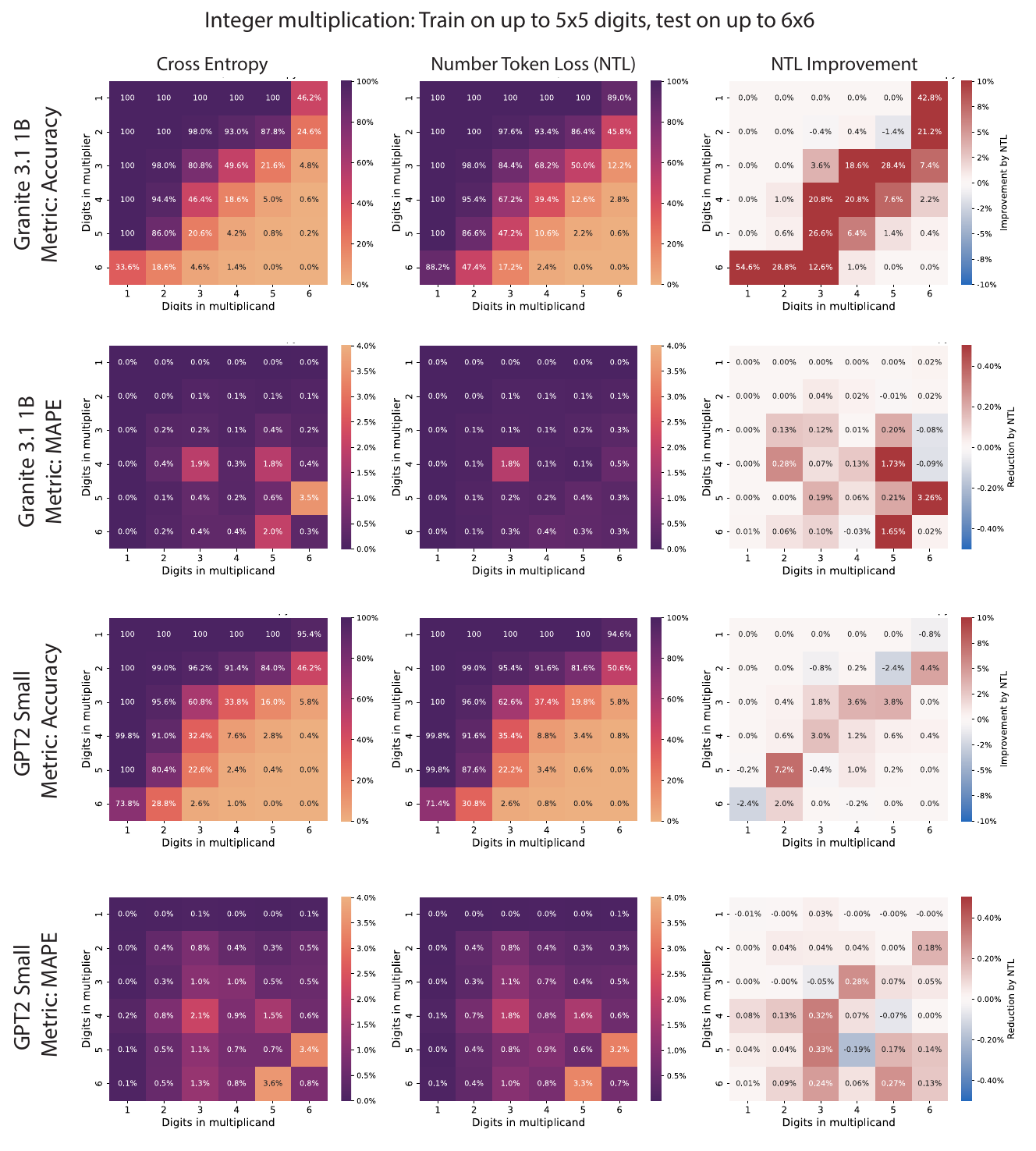}
    \caption{Performance improvement on the multiplication task using NTL, evaluated on GPT-2 and Granite 3.1.}
    \label{fig:multiplication_task}
\end{figure}

\begin{figure}[htb!]
    \centering
    \includegraphics[width=0.8\linewidth]{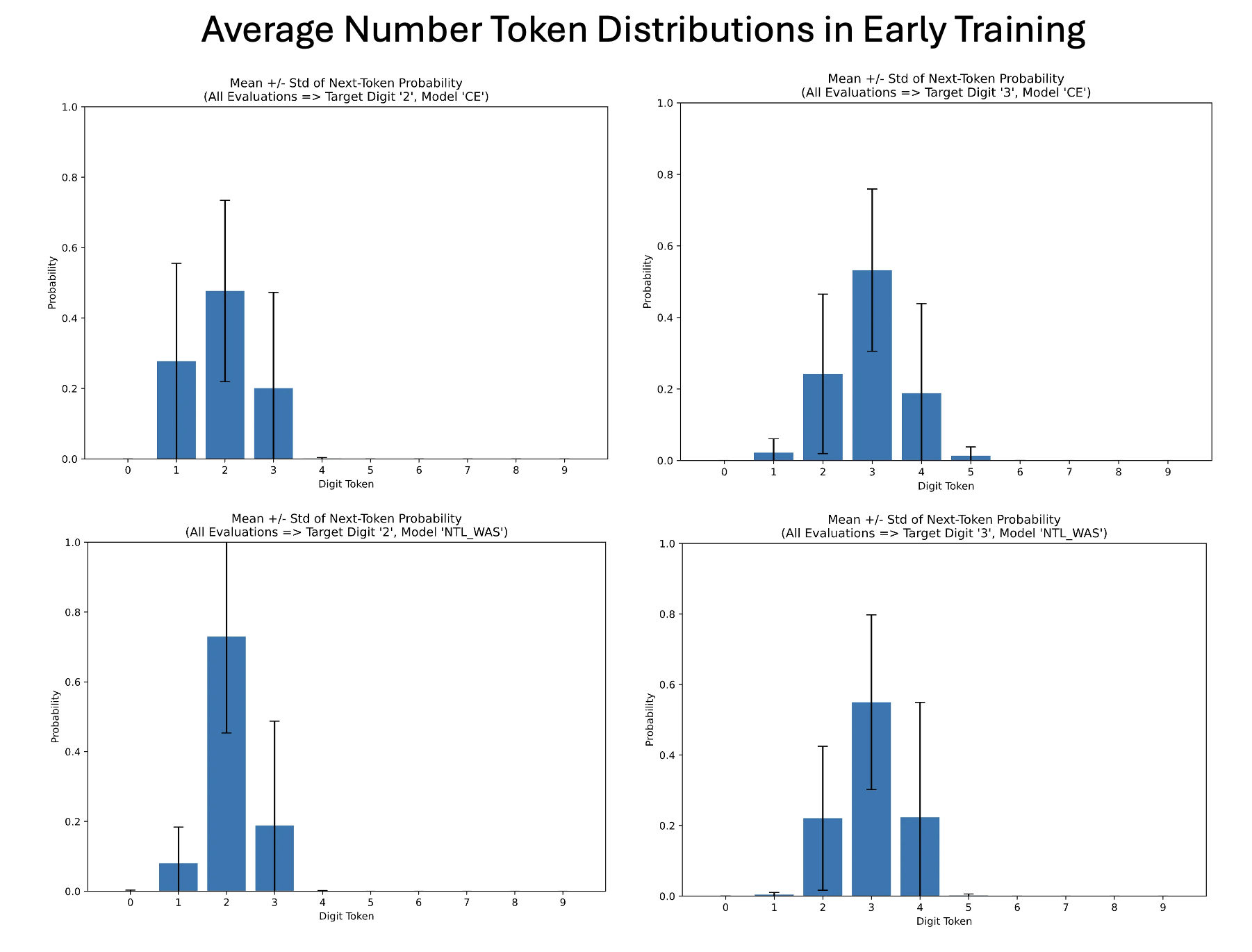}
    \caption{Average number token distributions for simple arithmetic tasks with target token $2$ and $3$ after training on the arithmetic task for 50k steps.}
    \label{fig:confidence}
\end{figure}

\FloatBarrier
\begin{example}[GSM8K sample -- NTL vs. CE]{ex:example1}
\footnotesize{
\textit{Question:}\\
Roberta wants to have a dinner party centered around soufflés. Each savory souffle calls for 8 eggs each and the dessert ones call for 6 eggs each. She wants to make 3 savory soufflés and 5 dessert soufflés for the party. How many eggs will she need?

\textit{Correct answer}:\\
The savory soufflés need 8 eggs each and she wants to make 3 of these so she needs 8*3 = 24 eggs The dessert soufflés need 6 eggs each and she wants to make 5 of these so she needs 6*5 = 30 eggs For the dinner party she will need 24+30 = 54 eggs in total \#\#\#\# 54

\textit{T5-CE}:\\
Roberta wants to make 3 savory soufflés that are 8 eggs each for a total of 3*8=24 eggs. She wants to make 5 dessert soufflés that are 6 eggs each for a total of 5*6=30 eggs. In total, she will need 24+30=62 eggs. \#\#\#\# 62

\textit{T5-NTL}:\\
Roberta wants to make 3 savory soufflés that use 8 eggs each so that’s 3*8 = 24 eggs She wants to make 5 dessert soufflés that use 6 eggs each so that’s 5*6 = 30 eggs All total, she needs 24 + 30 = 54 eggs \#\#\#\# 54

}
\end{example}

\end{document}